\documentclass[10pt,twocolumn,letterpaper]{article}

\usepackage[pagenumbers]{cvpr} 

\usepackage{graphicx}
\usepackage{amsmath}
\usepackage{amssymb}
\usepackage{booktabs}
\usepackage{appendix}

\usepackage{etoolbox}
\usepackage{amsfonts} 

\usepackage{amsmath}
\usepackage{amssymb,bm}
\usepackage{booktabs}
\usepackage{caption}
\usepackage{subcaption}

\usepackage{array}
\usepackage{booktabs}
\usepackage{caption}
\usepackage{multirow}
\usepackage{hhline}

\usepackage[colorlinks]{hyperref}

\def\cvprPaperID{*****} 
\def\confName{CVPR}
\def\confYear{2022}

\begin{document}

\title{ProtoPFormer: Concentrating on Prototypical Parts in Vision Transformers for Interpretable Image Recognition}

\makeatletter

\newcommand{\myfnsymbol}[1]{%
  \expandafter\@myfnsymbol\csname c@#1\endcsname
}
\newcommand{\@myfnsymbol}[1]{%
  \ifcase #1
  \or \TextOrMath{\textdagger}{\dagger}
  \fi
}



\renewcommand{\thefootnote}{\myfnsymbol{footnote}}

\author{
    Mengqi Xue\textsuperscript{$1,*$},
    Qihan Huang\textsuperscript{$2,*$},
    Haofei Zhang\textsuperscript{$2$},
    Jingwen Hu\textsuperscript{$2$}, \\
    Jie Song\textsuperscript{$2,3$},
    Mingli Song\textsuperscript{$1,2,3$}
    Canghong Jin\textsuperscript{$1,\dagger$}\\
    \textsuperscript{$1$}Hangzhou City University,
    \textsuperscript{$2$}Zhejiang University,
    \textsuperscript{$3$}ZJU-Bangsun Joint Research Center\\
     {\tt\small \{mqxue, jinch\}@hzcu.edu.cn,}\\
     {\tt\small \{qh.huang, haofeizhang, jw\_hu, sjie, brooksong\}@zju.edu.cn}\\
}

\maketitle

\footnotetext[1]{Corresponding author.}%
\footnotetext[2]{$^{*}$Equal contributions.}%

\begin{abstract}
Prototypical part network~(ProtoPNet) has drawn wide attention and boosted many follow-up studies due to its self-explanatory property for explainable artificial intelligence~(XAI). 
However, when directly applying ProtoPNet on vision transformer~(ViT) backbones, learned prototypes have a ``distraction'' problem: they have a relatively high probability of being activated by the background and pay less attention to the foreground. 
The powerful capability of modeling long-term dependency makes the transformer-based ProtoPNet hard to focus on prototypical parts, thus severely impairing its inherent interpretability. This paper proposes prototypical part transformer~(ProtoPFormer) for appropriately and effectively applying the prototype-based method with ViTs for interpretable image recognition. 
The proposed method introduces global and local prototypes for capturing and highlighting the representative holistic and partial features of targets according to the architectural characteristics of ViTs. The global prototypes are adopted to provide the global view of objects to guide local prototypes to concentrate on the foreground while eliminating the influence of the background.
Afterwards, local prototypes are explicitly supervised to concentrate on their respective prototypical visual parts, increasing the overall interpretability.
Extensive experiments demonstrate that our proposed global and local prototypes can mutually correct each other and jointly make final decisions, which faithfully and transparently reason the decision-making processes associatively from the whole and local perspectives, respectively. Moreover, ProtoPFormer consistently achieves superior performance and visualization results over the state-of-the-art~(SOTA) prototype-based baselines. Our code has been released at\textit{~\url{https://github.com/zju-vipa/ProtoPFormer}}.
\end{abstract}

\vspace{-1em}

\section{Introduction}

\begin{figure}[t]
\centering
    \includegraphics[width=\linewidth]{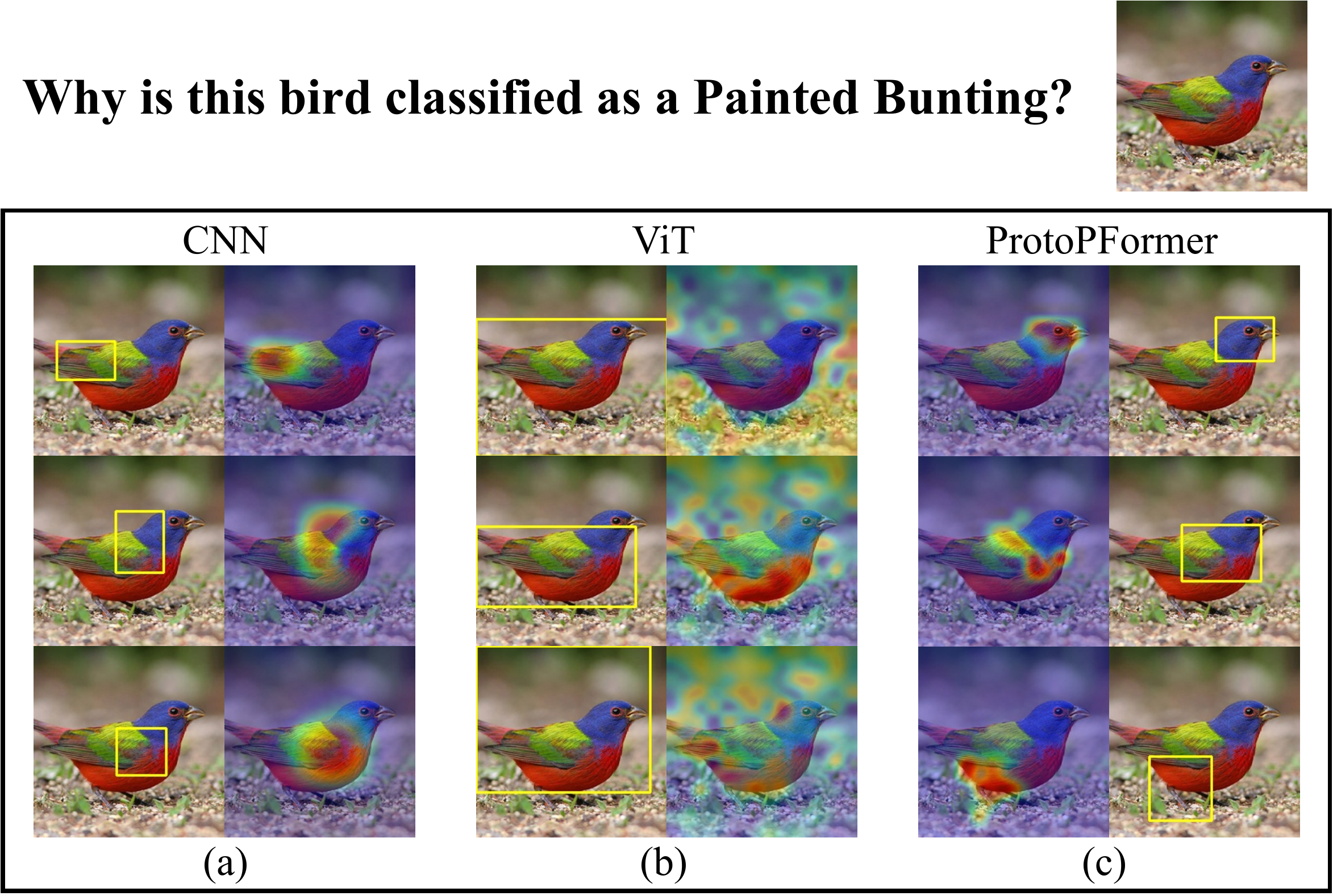}
\caption{Visual comparison of prototypes on an example image between a CNN-based ProtoPNet~(ResNet34~\cite{he2016resnet}) and a ViT-based ProtoPNet~(DeiT-Ti~\cite{touvron2021deit}), and our ProtoPFormer~(DeiT-Ti).}
\label{fig-1}
\end{figure}

The emergence of deep neural networks~(DNNs) has created unprecedented achievements in computing, thanks to their abilities to powerfully extract features and automatically mine recognition patterns in enormous data~\cite{lecun2015deep-dl}. However, the lack of transparency hinders DNNs' applications in areas that demand traceable and understandable decisions~\cite{buhrmester2021analysis-inter_review}.
To further explore the interpretability of DNNs, many researchers propose various approaches to promote the advancement in explainable artificial intelligence~(XAI)~\cite{buhrmester2021analysis-inter_review}. Among these methods, ProtoPNet~\cite{chen2019looks-protopnet}, inspired by the human vision system, has attracted increasing research interest and many follow-up studies with its self-explanatory property for XAI. Consider the example image in Figure~\ref{fig-1}, we can identify this sample as a Parakeet Auklet by comparing the features of its beak, wings, and feathers with existing bird species even without expertise.
Similar to this strategy, ProtoPNet aims to precisely perceive and recognize discriminative parts of objects with category-specific prototypes. As shown in Figure~\ref{fig-1}~(a), the highest activated prototypes capture features of the bird's head and wings. By making predictions through the linear combination of prototypes' similarity with image patches, ProtoPNet is inherently interpretable and can be analyzed by visualization when post-processing.

PorotoPNet and its variants are mainly developed on convolutional neural networks~(CNNs). While recent years vision transformers~(ViTs)~\cite{dosovitskiy2020vit} have been introduced into computer vision and challenged the domination of CNNs with their superior performance equipping the ability to model long-range dependency~\cite{liu2021swin,chen2022when}. Naturally, we want to see whether ViTs can benefit from prototype-based methods' self-explanatory property. 
Disappointingly, directly applying ProtoPNet with a ViT~(removing the class token) leads to a ``prototype distraction'' problem:  prototypes are prone to obtain high activation scores in the background and show scattered and fragmented activation scores in the foreground, as shown in~Figure~\ref{fig-1}~(b). The unsatisfactory visualization results violate the idea that makes prototypes point out vital visual evidence for each case. 
The lack of intrinsic inductive bias makes prototypes of ViTs focus less on prototypical parts and more on the long-range dependency.

To solve the aforementioned limitations, we propose \textit{prototypical part transformer}~(ProtoPFormer) to appropriately and effectively apply the prototype-based method with ViTs for interpretable image recognition in two steps, as shown in Figure~\ref{fig-1}~(c). ProtoPFormer proposes global and local prototypes to concentrate the ``distracted'' prototypes on the discriminative parts to build a self-interpretable ViT-backend model.
As reported by Raghu \etal ~\cite{raghu2021vision-do_vit_see_like_cnn}, the class token~(\ie, the global branch) of ViTs progressively aggregates information from all the image tokens and produces the high-level abstraction of targets; on the contrary, image tokens~(\ie, the local branch) remain strong similarities to their corresponding spatial locations in inputs. Hence the global and local prototypes are designed to compute similarity scores with output embeddings of the class token and image tokens, for capturing and highlighting holistic and partial features of targets and fully capitalizing on the built-in architectural characteristics of ViTs.
Next, we gradually perform a two-step concentration process to solve the ``distraction'' problem with the proposed two types of prototypes.
In the first step, the global prototypes perceive the whole features of targets and devise a foreground preserving~(FP) mask to guide the local branch to selectively keep foreground-related image tokens and eliminate the influence of the background. In the second step, A prototypical part concentration~(PPC) loss is designed to promote inter-prototype divergence and centralize scattered similarity scores, encouraging local prototypes further focus on diverse prototypical parts as visual explanations.
Finally, the predictions from the local and global branches are combined to make final decisions jointly. Our experiments have proved that combining the two types of high-level abstracted features can mutually correct each other’s decisions from their exclusive views. Moreover, extensive experiments have demonstrated that our ProtoPFormer not only enjoys superior performance and faithfully reasons the decision-making processes from both global and local perspectives, appropriately and efficiently resolving the limitations of previous prototype-based methods in ViTs.

We summarized our contributions as follows:
(a) We propose ProtoPFormer, tailored for ViT-backed prototype methods, for interpretable image recognition. (b) Based on the architectural characteristics of ViTs, ProtoPFormer introduces global and local prototypes to capture and highlight the holistic and partial features of target objects complementarily. (c) A two-step process is performed to progressively solve the ``prototype distraction'' problem and point out visual evidence associatively from global and local perspectives.
(d) Extensive experiments have demonstrated that ProtoPFormer can achieve superior performance and faithfully and transparently reason the decision-making processes, benefiting from the strategy of mutual correction and joint decision of global and local prototypes.



 

\section{Related Work}





\subsection{Interpretability with CNNs}

The interpretability of the inherent decision-making process of DNNs has become a grand challenge in computer vision. In general, previous works of model interpretability can be divided into two groups: \textit{self-interpretable models} and \textit{post-hoc analysis}. Self-interpretable models are elaborately designed neural networks that have transparent reasoning processes with regularization techniques~\cite{chen2016ad-hoc1,you2017ad-hoc2,subramanian2018ad-hoc3} or accountable components~\cite{zhang2018interpretable-cnn,zhang2019zqs1,zhang2018zqs2}. The post-hoc analysis methods focus on undermining interpretable information of a well-trained DNN with various techniques like feature visualization~\cite{molle2018post-hoc1-feature1,bychkov2018post-hoc1-feature2}, saliency analysis~\cite{zech2018post-hoc2-saliency1,chefer2021post-hoc2-saliency2} and gradients~\cite{selvaraju2017post-hoc2-saliency3-grad1,singla2019post-hoc2-saliency3-grad2,smilkov2017post-hoc2-saliency3-grad3}.
ProtoPNet~\cite{chen2019looks-protopnet} combines the characteristics of both schools with faithfully reasoning the decision-making process by the linear combination of the prototype's similarity scores and visualizing the importance of discriminative parts as post-hoc analysis.
Many follow-up studies extend ProtoPNet to medical image processing, explanatory debugging, etc~\cite{singh2021these-NP_ProtoPNet,keswani2022proto2proto,nauta2021neural-prototree,donnelly2022deformable-protopnet,wang2021interpretable-proto_TesNet,bontempelli2022concept-ProtoPDebug}. While these CNN-based methods tend to obtain unsatisfactory results both on classification accuracies and visualization of prototypes when directly applying to ViTs~\cite{dosovitskiy2020vit}. Hence we propose ProtoPFormer to solve these limitations.

\begin{figure*}[!t]
\centering
    \includegraphics[width=0.96\linewidth]{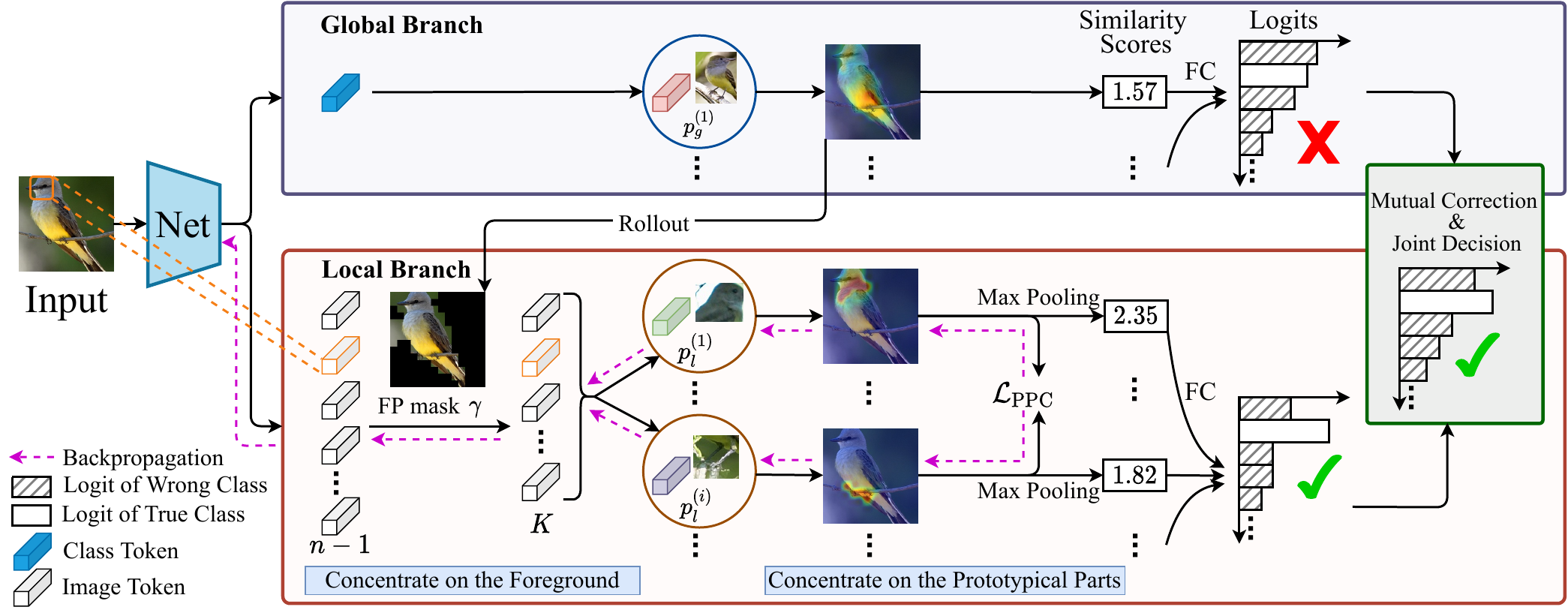}
\caption{Illustration of ProtoPFormer for image recognition interpretation. The global branch provides guidance for the local branch with the FP mask. The strategy of mutual correction and joint decision makes them contribute complementarily to final predictions, capitalizing on the built-in architectures in ViTs. The loss propagation of $\mathcal{L}_{\mathrm{CE}}$ is omitted for simplicity.}

\label{fig-method}
\end{figure*}



\subsection{Interpretability with ViTs}
Transformer\cite{vaswani2017attention-all_you_need} is introduced into computer vision filed and has achieved impressive success~\cite{touvron2021deit,liu2021swin}. With the wide applications of ViTs, some approaches are proposed to explore their interpretability. The most intuitive way is to analyze the attention weights ~\cite{vaswani2017attention-all_you_need,abnar2020quantifying-rollout}. Nonetheless, this simple assumption may not be a fail-safe indicator~\cite{serrano2019attention-interpretable_nlp}. For ViTs, some reasons the decision-making process via gradients~\cite{gao2021ts-tscam,gupta2022vitol,chen2022lctr}, attributions~\cite{chefer2021transformer-interpretability_attribution,yuan2021explaining-Markov_chain} and redundancy reduction~\cite{pan2021ia-IA_RED}. While these methods can only attend to the global features, leaving out discriminative parts.
In particular, ViT-NeT~\cite{kim2022vit-vit_net} and ConceptTransformer~\cite{rigotti2021attention-ConceptTransformer} include visual prototypes/concepts into ViTs as visual explanations. While  ViT-NeT merely adopts ViTs as backbones to extract features. The interpretability mainly comes from the neural tree that also introduces many parameters and ignores the architectural character of ViTs. 
ConceptTransformer adds user-defined concepts~(\eg, attribute annotations) to enforce an additive relation between token embeddings and concepts. The user-defined concepts require expensive and time-consuming human labeling.
Comparatively, ProtoPFormer is designed precisely for ViTs according to the self-attention mechanism, the class token, and the image tokens with category-specific prototypes that can be automatically learned when training. Furthermore, global and local branches can mutually correct each other and jointly make the final decision, capitalizing on the built-in class token and image tokens in ViTs.

\vspace{-1em}
\section{Preliminaries}

\noindent\textbf{ProtoPNets.}
A typical ProtoPNet is composed of three sequential modules: (1) a backbone network maps an input image to a sequence\footnotemark of visual tokens~(embedding vectors) $X\in\mathbb{R}^{n\times d}$, where $n$ is the length of the visual sequence and $d$ is the embedding dimension;
(2) a prototype layer $\mathrm{Proto}(X)$ transforms $X$ into a similarity score vector $s\in\mathbb{R}^{m}$, where $m$ denotes the number of learnable prototypes;
(3) a fully connected~(FC) layer $\mathrm{FC}(s)$ makes the prediction with $s$.
In the prototype layer, particularly, the $i$-th similarity score $s_i$ is the max pooled value from the similarity map $S_i=\mathrm{Sim}(X, p^{(i)})$, where $p^{(i)}\in\mathcal{P}$ is the $i$-th prototype and $\mathrm{Sim}(\cdot, \cdot)$ computes the similarity between the given prototype $p_i$ and all visual tokens, defined in~\cite{chen2019looks-protopnet}.
Generally, ProtoPNets assign $k$ prototypes equally for each class and therefore $m=kC$~($C$ is the number of classes).

\footnotetext{The feature map~(shape $h\times w\times d$) output from CNNs can be flattened as a visual sequence with shape $hw\times d$.}

\noindent\textbf{Vision Transformers.}
This paper focuses on ViTs adopting the attention mechanism as the original Transformer~\cite{vaswani2017attention-all_you_need}. ViTs firstly embed disjoint image patches as a sequence of image tokens.
Then they are appended with a class token and fed into multiple encoder layers composed of a multi-head self-attention~(MHSA) module and an MLP block.
Given a visual sequence $X\in\mathbb{R}^{n\times d}$, according to~\cite{vaswani2017attention-all_you_need} the MHSA layer can be
rewritten as follows
\begin{equation}
    \mathrm{MHSA}(X) = \sum_{h=1}^H A_h X W_h \text{,}
    \label{eq:mhsa}
\end{equation}
where $A_h=\mathrm{softmax}(\Psi_h)$ is the normalized self-attention matrix by row-wise softmax of head $h$ ($\Psi_h \in\mathbb{R}^{n\times n}$), $H$ is head number, and $W_h\in\mathbb{R}^{d\times d}$ is the linear projection matrix.
In particular, the global information is gradually aggregated to the class token solely for the final classification by each MHSA layer, 



\noindent\textbf{ViT-backed ProtoPNets.}
To directly implement a ViT as the backbone, we remove the class token from the sequence and feed the remaining tokens to the following prototype layer.
However, unlike CNNs, which are stacked with local perceptron units, \eg, convolutional and pooling layers, the MHSA mechanism aggregates global information to every visual token, which is consequently incompatible with prototypes representing local visual parts.
Furthermore, as extensive experiments demonstrates in Table~\ref{tab-1} and Figure~\ref{fig-4}, ViT-backed ProtoPNet and its variants show notably performance degeneration compared with CNN-backed ProtoPNets.
This phenomena can be explained by visualizing the learned prototypes shown in Figure~\ref{fig-4}, revealing a ``prototype distraction'' problem that the tokens with high similarity scores are distributed irregularly and spread the whole similarity map, including background positions.
As a result, not only the classification accuracy but also the interpretability coming with the semantic meaning of prototypes suffers from the incompatibility between the long-range perception of ViTs and the local visual dependency of ProtoPNets.

\section{ProtoPFormer}

To tackle the abovementioned problem, we propose explicitly constraining the activated similarity map in the local branch so that every prototype is enforced to represent an individual visual part as intended.
Moreover, employing the global information aggregated to the class token, the global branch is proposed to model inter-class and intra-class differences contributing complementarily to final predictions along with the local prototype branch.

The overall architecture is illustrated in~Figure~\ref{fig-method}, in which a class token $t_c\in\mathbb{R}^{1\times d}$ and a feature sequence $X_f\in\mathbb{R}^{(n - 1)\times d}$ are split from the visual sequence $X=[t_c, X_f]$ and fed into a global prototype branch and a local prototype branch respectively.
Similar to the CNN-backed ProtoPNets settings, the global branch has $m_g$ learnable prototypes $\mathcal{P}_g=\{p_g^{(1)}, \ldots, p_g^{(m_g)}\}$~($m_g^c$ for each class), and the local branch has $m_l$ learnable prototypes $\mathcal{P}_l=\{p_l^{(1)}, \ldots, p_l^{(m_l)}\}$~($m_l^c$ for each class).
The predicted logits of the two branches are weighted summed for generating the complementary result, formulated as
\begin{equation}
    z_c = \lambda_g z_g + \lambda_l z_l \text{,}
\end{equation}
where $z_g$ and $z_l$ are outputs from the FC classification layer of global and local branches in respective, and $\lambda_{\{g,l\}}$ are weighted coefficients for each branch.


\subsection{Concentration on the Foreground}
The first step is to concentrate local prototypes on the foreground and eliminate the influence of the background via an adaptive binary mask, named foreground preserving~(FP) mask, to selectively preserve foreground-related image tokens and screen background-related image tokens.
The global branch provide guidance to the local branch via the FP mask, capitalizing on the built-in class token and image tokens in ViTs.
As shown in Figure~\ref{fig-method}, we use the rollout method~\cite{abnar2020quantifying-rollout} to generate the FP mask from the rollout matrix of the class token.
For a ViT model, the attention rollout matrix at the $l$-th layer~($l \geq 1$) is defined recursively as $\tilde{\mathcal{A}}_r^{(l)} = \mathcal{A}^{(l)} \tilde{\mathcal{A}}_r^{(l-1)}$,
where $\mathcal{A}^{(l)}$ is computed based on the attention matrices $A_h^{(l)}$ at this layer:
\begin{equation}
    \mathcal{A}^{(l)} = I_n + \frac{1}{H} \sum_{h=1}^H A_h^{(l)} \text{.}
\end{equation}
The initial attention rollout matrix $\mathcal{A}^{(0)}$ is predefined as the identity matrix $I_n$.
As mentioned in~\cite{abnar2020quantifying-rollout}, $\tilde{\mathcal{A}}_{1,i}$ is the influence score of the $i$-th token to the class token, which help us distinguish how likely it could be as a foreground token as visualization in Figure~\ref{fig-method}.

Given a backbone ViT with $L$ encoder layer, we extract the rollout matrix at the $(L-1)$-th layer for filtering out background tokens.
Let $\hat{a}_c\in\mathbb{R}^{n-1}$ be the rollout attention values to the class token, we only preserve top-$K$ foreground tokens for computing the $L$-th encoder layer.
More specifically, let $\gamma\in\{0, 1\}^{n-1}$ denote a binary foreground preserving~(FP) mask ($\gamma_i = 1$ represents that token $i$ is a preserved foreground token).
To remove selected background tokens, we modify the softmax normalization applied to $\Psi$ in~Equation~\ref{eq:mhsa} as
\begin{equation}
    A_{i,j} = \frac{\gamma_j \exp(A_{i,j})}{\sum_{k=1}^{n-1} \gamma_k \exp(A_{i,k})} \text{.}
    \label{eq:modify}
\end{equation}
Equation~\ref{eq:modify} cuts off the connection between the background and the foreground, thus avoiding changing information at this step. In the following prototype layer, 
only the similarity scores generated from the foreground-related tokens are preserved for further class prediction.
Generally, this step tentatively concentrates local prototypes in the foreground with the proposed FP mask.


\subsection{Concentration on Prototypical Parts}
With selected foreground tokens, the local prototypes can be forced to concentrate on heterogeneous visual parts with explicit supervision.
For achieving such a purpose, we model the similarity map~(reshaped to a two-dimension array like image patches) with regard to each local prototype as a bivariate Gaussian function:
\begin{equation}
    \mathrm{Gaussian}(x|\mu, \Sigma) = \frac{1}{2\pi |\Sigma|^{\frac{1}{2}}} e^{-\frac{1}{2} (x-\mu)^\top \Sigma^{-1} (x-\mu)} \text{,}
\end{equation}
where $x\in\mathbb{R}^2$ represents the position on the similarity map; $\mu\in\mathbb{R}^2$ and $\Sigma\in\mathbb{R}^{2\times 2}$ are the parameters controlling the center position and dispersion in respective.
On the one hand, by minimizing the eigenvalues of $\Sigma$, we are able to dissolve the distraction problem.
On the other hand, to promote the divergency of prototypes, we can supervise the prototypes by pushing the centers away from each other.

\noindent\textbf{Gaussian Fitting.}
When achieving $N$ data points $D = \{(x_i, s_i)\}_{i=1}^{N}$ ($x_i$ represent the position of the similarity value $s_i$ on the heatmap $S$) along with the FP mask $\gamma$, we are able to estimate the parameters of the Gaussian model $\mathrm{Gaussian}(\cdot|\mu, \Sigma)$.
Precisely, by removing background tokens, $\mu$ and $\Sigma$ can be estimated by
\begin{equation}
\left\{
    \begin{aligned}
        \hat{\mu} &= \frac{\sum_{i=1}^N \gamma_i s_i x_i }{\sum_{i=1}^N \gamma_i s_i} \text{,} \\
        \hat{\Sigma} &= \frac{\sum_{i=1}^N \gamma_i s_i (x_i - \mu) (x_i - \mu)^\top}{\sum_{i=1}^N \gamma_i s_i - 1} \text{.}
    \end{aligned}
\right. 
\end{equation}

Then we propose a prototypical part concentration~(PPC) loss to make local prototypes concentrate on different and centralized representative parts for each class.
By constraining $\mathrm{tr}(\hat{\Sigma})$, the trace of $\hat{\Sigma}$, the PPC loss minimizes the sum of eigenvalues of $\hat{\Sigma}$. In the meantime, this loss also encourages prototypes belonging to the same class to have diverse $\hat{\mu}$.
The PPC loss can be computed as $\mathcal{L}_{\mathrm{PPC}}=\lambda_{\mu}\mathcal{L}_{\mathrm{PPC}}^{\mu}+\lambda_{\sigma}\mathcal{L}_{\mathrm{PPC}}^{\sigma}$, where $\mathcal{L}_{\mathrm{PPC}}^{\mu}$ is defined as
\begin{equation}
    \mathcal{L}_{\mathrm{PPC}}^{\mu}= \frac{1}{m_c^lm_c^l}\sum_{i \neq j}\mathrm{max}(t_{\mu} - \|\hat{\mu}_i^{c} - \hat{\mu}_j^{c}\|^2, 0) \text{,}
\end{equation}
and $\mathcal{L}_{\mathrm{PPC}}^{\sigma}$ can be written as
\begin{equation}
    \mathcal{L}_{\mathrm{PPC}}^{\sigma} = \mathrm{tr}\left(\mathrm{max}(0, \hat{\Sigma} - t_\sigma)\right) \text{.}
\end{equation}
Here $t_{\mu}$ and $t_{\sigma}$ are two predefined thresholds to guarantee that the PPC loss only penalizes too close center coordinates and encourages small covariance values for concentrating  $c$-class private local prototypes to learn and make decisions on different distinctive visual concepts. And $\lambda_{\mu}$ and $\lambda_{\sigma}$ represent their factors. The final optimization objection of ProtoPFormer is to minimize $\mathcal{L} = \mathcal{L}_{\mathrm{CE}} + \mathcal{L}_{\mathrm{PPC}}$, $\mathcal{L}_{\mathrm{CE}}$ is the conventional cross-entropy loss.

\section{Experiments}

\subsection{Experimental Settings}

\noindent\textbf{Datasets.}
We conduct experiments on three widely-used datasets including CUB~\cite{399-cub}, Dogs~\cite{KhoslaYaoJayadevaprakashFeiFei_FGVC2011-dogs} and Cars~\cite{krause20133d-stanfordcars}. The CUB dataset contains 5994/5794 training/test images over 200 bird species. The Dogs dataset is a fine-grained classification dataset with 120 dog breeds. The Cars dataset consists of 16,185 vehicle pictures of 196 classes of cars. All images are resized to $224\times224$ pixels without cropping.

\noindent\textbf{Backbones.}
Three popular transformers, DeiT-Ti~\cite{touvron2021deit}, DeiT-S~\cite{touvron2021deit}, and CaiT-XXS-24~\cite{touvron2021going-cait}, are adopted as the ViT backbones in our experiments, initialized with the official pre-trained weights on ImageNet-1k~\cite{deng2009imagenet}.

\noindent\textbf{Parameters.}
All models are trained for 200 epochs with AdamW optimizer~\cite{loshchilov2017decoupled-adamw} and cosine LR scheduler, following the training strategies used in official codes of DeiT. Grid search is used to tune all involved hyperparameters of our method and baselines.
The weighted coefficients $\lambda_{\mu}, \lambda_{\sigma}$ are set to 0.5, 0.1, and the two thresholds $t_{\mu}, t_{\sigma}$ are 2 and 1.
We use 10 local prototypes for all datasets, and 10, 5, and 5 global prototypes for CUB, Dogs and Cars datasets, respectively. $K$ is 81 for CUB and Dogs, and 121 for Cars. FC layers are non-trainable in ProtoPFormer.

\begin{table*}
\renewcommand\arraystretch{1.15}
\centering
\setlength{\tabcolsep}{0.8mm}{
\begin{tabular}{c|*3{c}|*3{c}|*3{c}}
  \toprule
\multirow{2}*{\textbf{Method}} & \multicolumn{3}{c|}{\textbf{DeiT-Ti}}  & \multicolumn{3}{c|}{\textbf{DeiT-S}}  & \multicolumn{3}{c}{\textbf{CaiT-XXS-24}} \\

\cline{2-10}

& \textbf{CUB} & \textbf{Dogs} & \textbf{Cars} & \textbf{CUB} & \textbf{Dogs} & \textbf{Cars} & \textbf{CUB} & \textbf{Dogs} & \textbf{Cars} \\



  \toprule
  
\textbf{Base} & 80.57~\textit{\tiny5.56M} & 81.05~\textit{\tiny5.55M} & 86.21~\textit{\tiny5.56M} & 84.28~\textit{\tiny21.74M} & 89.00~\textit{\tiny21.71M} & 90.06~\textit{\tiny21.74M} & 83.95~\textit{\tiny11.80M} & 85.62~\textit{\tiny11.79M} & 90.19~\textit{\tiny11.80M} \\
\textbf{ProtoPNet} & 81.36~\textit{\tiny5.95M} & 81.47~\textit{\tiny5.79M} & 86.84~\textit{\tiny5.94M} & 84.04~\textit{\tiny22.12M} & 86.85~\textit{\tiny21.97M} & 88.21~\textit{\tiny22.12M} & 84.02~\textit{\tiny12.18M} & 84.62~\textit{\tiny12.03M} & 88.87~\textit{\tiny12.18M} \\
\textbf{ProtoTree} & 68.50~\textit{\tiny5.70M} & 68.46~\textit{\tiny5.70M} & 70.02~\textit{\tiny5.70M} & 70.57~\textit{\tiny21.89M} & 72.73~\textit{\tiny21.89M} & 74.95~\textit{\tiny21.89M} & 72.33~\textit{\tiny11.94M} & 73.69~\textit{\tiny11.94M} & 73.15~\textit{\tiny11.94M} \\
\textbf{TesNet} & 77.72~\textit{\tiny6.38M} & 76.54~\textit{\tiny5.97M} & 84.69~\textit{\tiny6.36M} & 81.36~\textit{\tiny22.56M} & 75.08~\textit{\tiny22.15M} & 87.31~\textit{\tiny22.54M} & 81.52~\textit{\tiny12.62M} & 77.01~\textit{\tiny12.21M} & 88.12~\textit{\tiny12.6M} \\
\textbf{Def. ProtoPNet} & 75.79~\textit{\tiny7.86M} & 79.26~\textit{\tiny6.99M} & 82.01~\textit{\tiny7.82M} & 79.53~\textit{\tiny25.93M} & 79.59~\textit{\tiny24.45M} & 87.42~\textit{\tiny25.86M} & 81.09~\textit{\tiny14.10M} & 80.67~\textit{\tiny13.23M} & 87.51~\textit{\tiny14.05M} \\
\textbf{CT} & 74.71~\textit{\tiny5.84M} & N/A & N/A & 79.74~\textit{\tiny22.75M} & N/A & N/A & 78.81~\textit{\tiny12.08M} & N/A & N/A \\
\textbf{ViT-Net} & 81.98~\textbf{\textit{\tiny10.42M}} & 80.96~{\textbf{\textit{\tiny10.08M}}} & 88.41~\textbf{\textit{\tiny10.41M}} & 84.26~{\textbf{\textit{\tiny26.66M}}} & 88.21~\textbf{\textit{\tiny26.30M}} & \textbf{91.34}~\textbf{\textit{\tiny26.64M}} & 84.51~\textbf{\textit{\tiny16.32M}} & 84.67~\textbf{\textit{\tiny16.32M}} & \textbf{91.54}~\textbf{\textit{\tiny16.65M}}\\
\hline
\textbf{ProtoPFormer} & \textbf{82.26}~\textit{\tiny6.33M} & \textbf{82.20}~\textit{\tiny5.91M} & \textbf{88.48}~\textit{\tiny6.13M} & \textbf{84.85}~\textit{\tiny22.50M} & \textbf{89.97}~\textit{\tiny22.09M} & 90.86~\textit{\tiny22.30M} & \textbf{84.79}~\textit{\tiny12.57M} & \textbf{86.26}~\textit{\tiny12.15M} & 91.04~\textit{\tiny12.36M} \\

  \bottomrule
  \end{tabular}}
\caption{The acc@1 performance comparison~(\%) between seven SOTA baselines and our ProtoPFormer on three datasets averaged over six runs. Big bold fonts are used to indicate the best accuracy. Tiny italic fonts report parameter numbers of each method, and the biggest model sizes are represented in bold type.}
\label{tab-1}

\end{table*}

\noindent\textbf{Baselines.}
We compare the proposed ProtoPFormer with the classic and state-of-the-art~(SOTA) prototype-based approaches. 
(1) \textbf{Base} represents the vanilla ViT model, serving as the non-interpretable counterpart of our method. 
(2) \textbf{ProtoPNet}~\cite{chen2019looks-protopnet} is the first work that interprets DNNs' decisions through a linear combination of similarity scores of prototypes. (3) \textbf{ProtoTree}~\cite{nauta2021neural-prototree} combines prototypes with decision trees for hierarchical reasoning. (4) \textbf{TesNet}~\cite{wang2021interpretable-proto_TesNet} introduces a transparent embedding space with class-aware basis concepts. 
(5) \textbf{Def.ProtoPNet}~\cite{donnelly2022deformable-protopnet} designs spatially flexible prototypes for handling images with pose variations. 
(6) \textbf{CT}~\cite{rigotti2021attention-ConceptTransformer} stands for ConceptTransformer which utilizes attributes as visual concepts. This method cannot be applied to Dogs and Cars which lack visual attribute labels.
(7) \textbf{ViT-Net}~\cite{kim2022vit-vit_net} integrates ViTs and trainable neural trees based on ProtoTree, which only employs ViTs as feature extractors without fully exploiting their architectural characteristics. For fair comparison, we rerun all baselines with three ViT backbones and use grid search to turn their hyperparameters. Please refer to the supplementary material for more detailed settings like hyperparameters of baselines and experimental implementations.

\begin{figure}[h]
\centering
    \includegraphics[width=\linewidth]{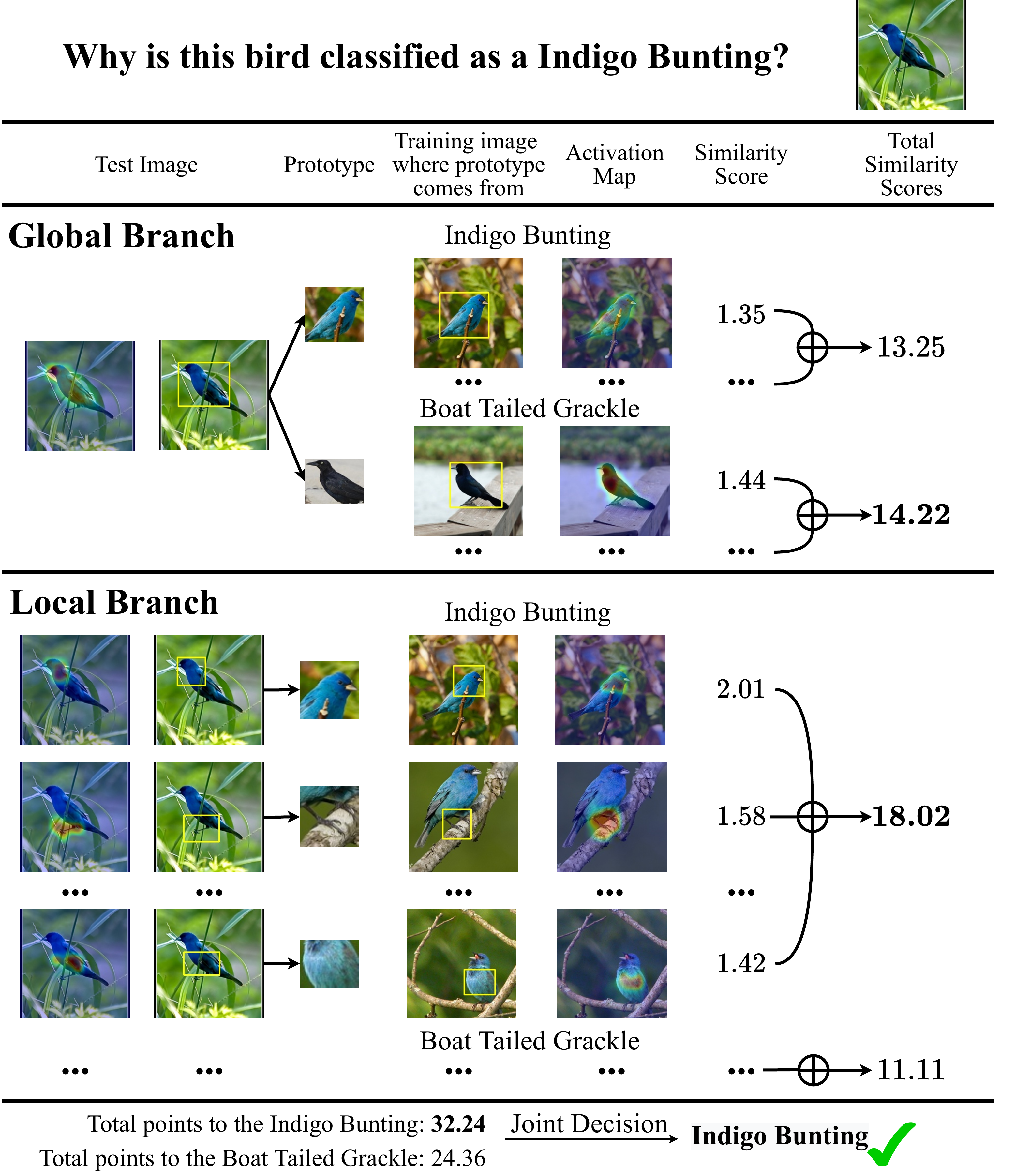}
\caption{The reasoning process of our ProtoPFomer in classifying the species of a bird with DeiT-Ti, where $\bm{\oplus}$ denotes summation of similarity scores.}
\label{fig-3}
\end{figure}

\begin{figure*}[htb]
\centering
    \includegraphics[width=\linewidth]{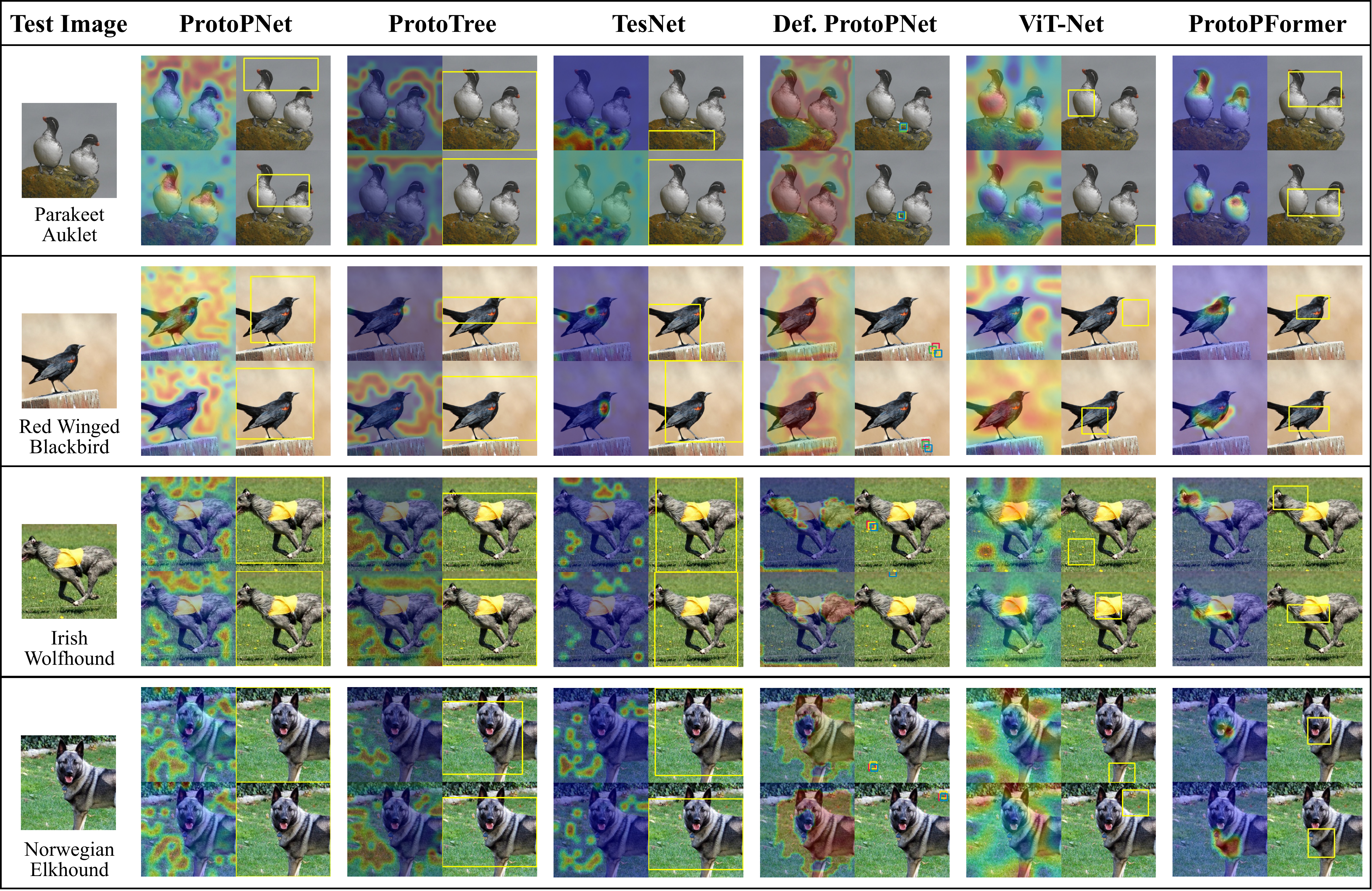}
\caption{Visual demonstration of the two most activated local prototypes in heat maps and bounding boxes on example images~(randomly chosen from the CUB and Dog datasets) of five prototype-based baselines and ProtoPFormer with DeiT-S.}
\label{fig-4}
\end{figure*}

\begin{figure}[h]
\centering
    \includegraphics[width=\linewidth]{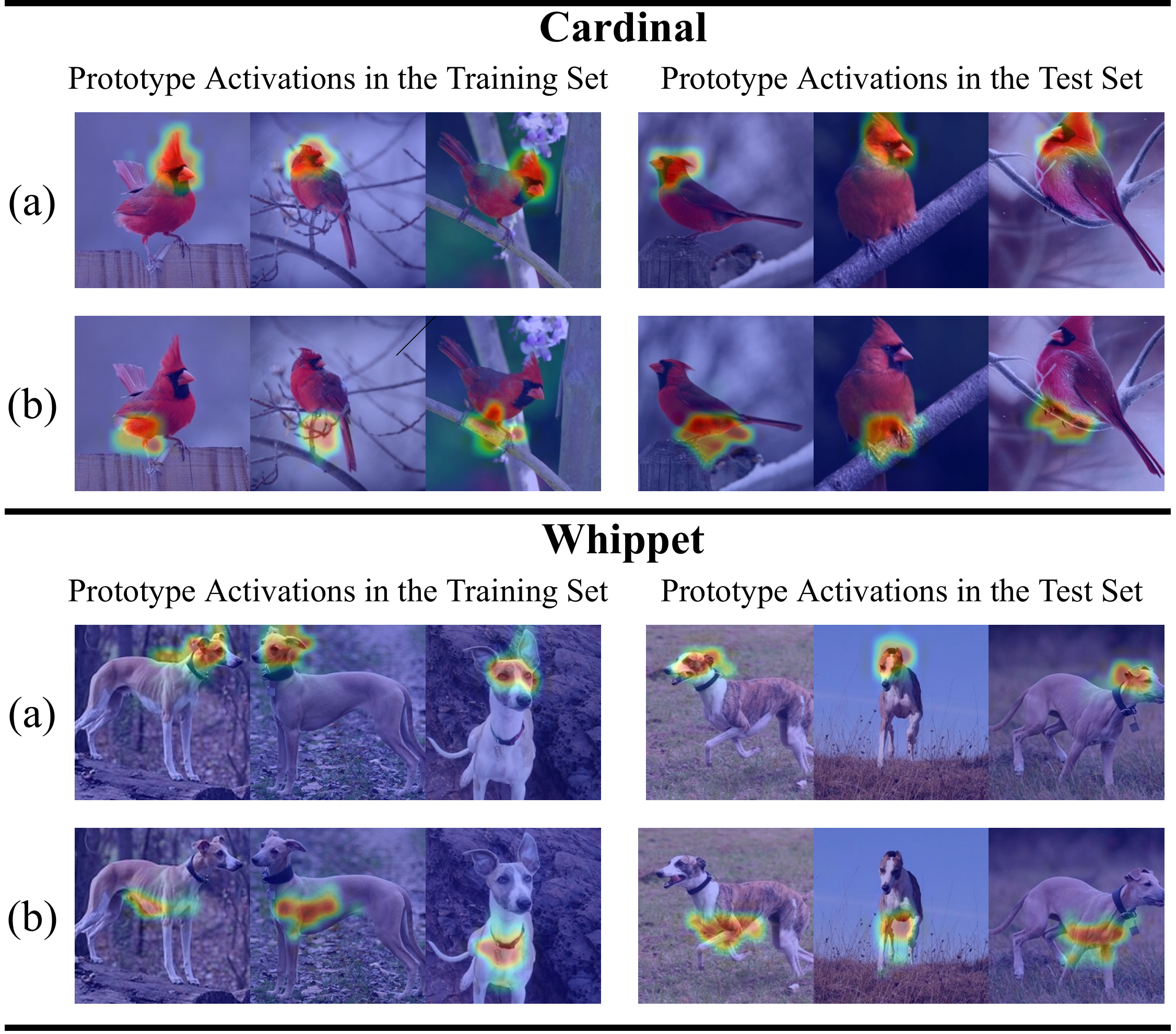}
\caption{Heat maps of the same local prototypes on different examples from the training and test set with DeiT-S.}
\label{fig-5}
\end{figure}

\subsection{Performance Comparison}

In Table~\ref{tab-1}, we report the top-1 accuracy and parameter numbers of ProtoPFormer and our competitors on involved datasets with three ViT backbones. Generally, it can be observed that our proposed method consistently achieves superior performance compared to baseline methods and economizing parameters. Specifically, Base, as the non-interpretable counterpart, steadily shows upper-middle performance than other self-interpretable baselines, which implies that previous SOTA prototype-based methods are confronted with the trade-off between accuracy and transparency for ViT-backed models.
ProtoPNet, TesNet, and Def.ProtoPNet suffer from varying degrees of performance degradation with transformer backbones on three datasets, indicating that these approaches, designed based on CNN models, are unsuited for learning prototypical parts from token embeddings in transformers. CT obtains worse results than our method with visual concepts defined by user-defined attributes on CUB, which require time-consuming labeling and rely on human judgment. In contrast, our prototypes are automatically learned along with the training process, faithfully reflecting the discriminative information for the decisions of ViTs.
ViT-Net has comparable accuracy with our method but introduces significant extra parameters with the neural trees.
Comparatively, ProtoPFormer solely adds small additional parameters, the global prototypes, to achieve the necessary improvement for ViTs. In summary, extensive experiments have verified that our ProtoPFormer outperforms seven baselines on adopted datasets with three ViTs, meanwhile avoiding increasing too many overheads.




\subsection{Visualization Analysis}

\noindent\textbf{Reasoning Process.}
Figure~\ref{fig-3} shows a typical reasoning process of ProtoPFormer. The global and local branches make complementary predictions contributing to the final decision. Prototypes of each branch compute corresponding similarity scores with token embeddings and produce the final points with linearly combined scores. In this case, when classifying an Indigo Bunting, global prototypes mistakenly have a slightly larger response with the holistic features of the Boat Tailed Grackle class than Indigo Bunting. Meanwhile, local prototypes successfully discover that the test bird reveals many close parallels with the Indigo-Bunting's exclusive prototypes: the indigo features covered the bird's head, belly, and black feet. Pointing out prototypical visual evidence helps local prototypes make sound judgments and correct the global branch. With the mutual correction and joint decision strategy, global and local prototypes capture holistic and partial features of target objects in a complementary way, benefiting the final decisions of ProtoPFormer.


\noindent\textbf{Visual Comparison.}
In this subsection, we analyze the interpretability of ProtoPFormer through visualizing local prototypes after training and compare the results with five baselines, illustrated in Figure~\ref{fig-4}. 
The bounding box covers the top 5$\%$ similarity scores in the same map, and heat maps are generated by up-sampling and mapping the activation maps to the pixel space, following the same visualization process in ProtoPNet.
We can observe that previous prototype-based competitors obtain unsatisfied visualization results: similarity scores are distributed irregularly and spread throughout the whole similarity map while paying excessive attention on the background positions. For example, two prototypes of ProtoPNet show high similarity with the background when classifying the Red Winged Blackbird, which dramatically impairs its self-explanatory. 
Nevertheless, ViT-Net and Def.ProtoPNet train prototypes to capture features of the greensward for identifying two dog breeds, as dogs often appear accompanied by the grass. We ascribe this phenomenon to the vulnerability of previous CNN-based baselines to ViT backbones, which mistakenly learn the misleading information related but not congruent with objects and ignore targeting cues. By contrast, ProtoPFormer preciously captures diverse discriminative prototypical parts with the two-step concentration. Notably, the high activated prototypes of ProtoPFormer show significant activations with a central tendency, \eg, the birds' head and under-surface and the dogs' head and belly. 
Moreover, heat maps of the two most activated local prototypes (a) and (b) on different examples from the training and test set in heat maps with are provided in Figure~\ref{fig-5}. Our ProtoPFomer can learn prototypes that attend to diverse visual evidence matches with human vision, \eg, the Cardinal's head and feet, the Whippet's head and belly, in each case. Global prototypes mainly capture the holistic features of targets with pose variations and different perspectives. Please refer to the supplement for more visualization analyses like visualization of global prototypes and Gaussian function shapes.







\begin{figure}[t]
    \centering
    \subcaptionbox{\normalsize {CUB}}{%
        \includegraphics[width=0.33\linewidth]{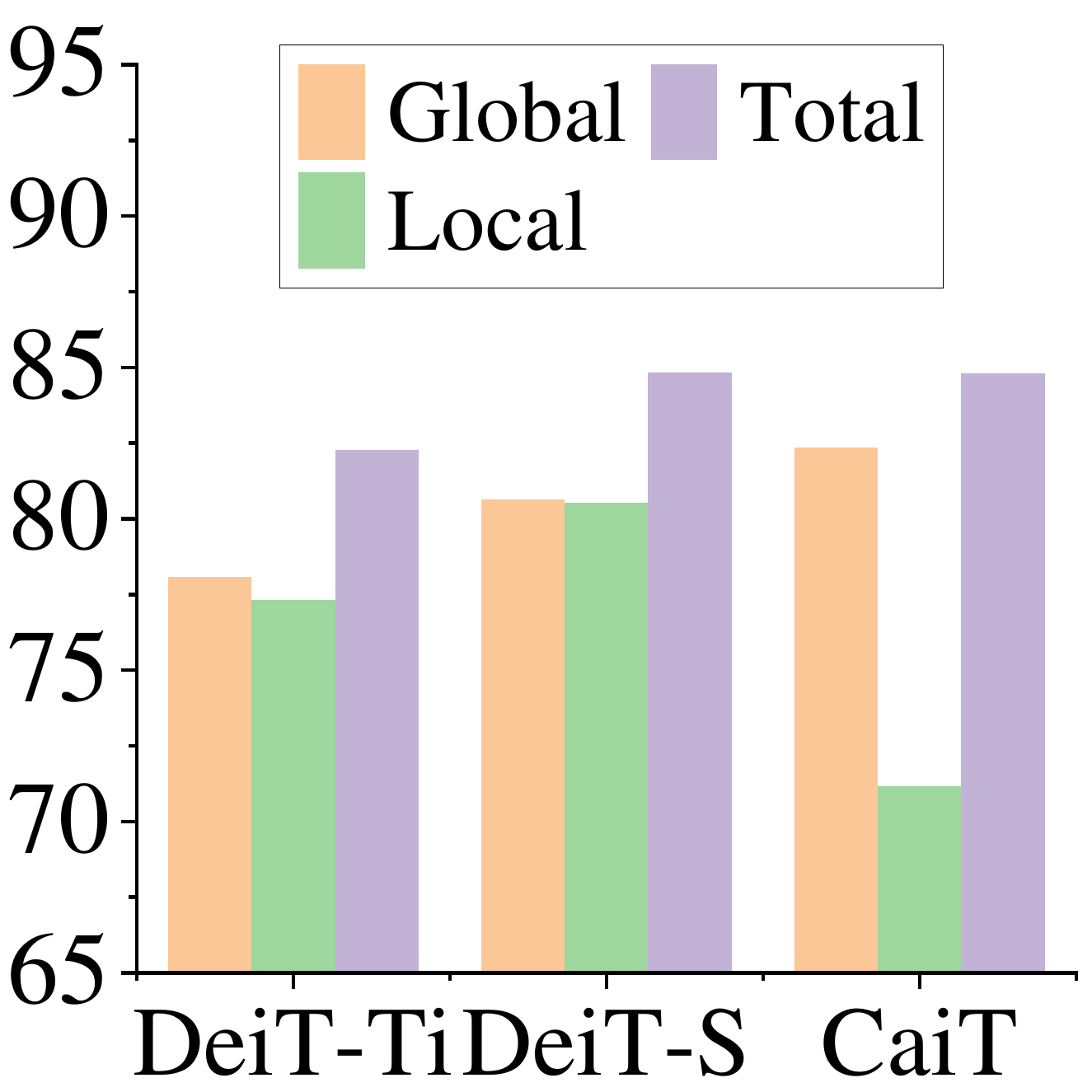}%
    }%
    \subcaptionbox{\normalsize {Dogs}}{%
        \includegraphics[width=0.33\linewidth]{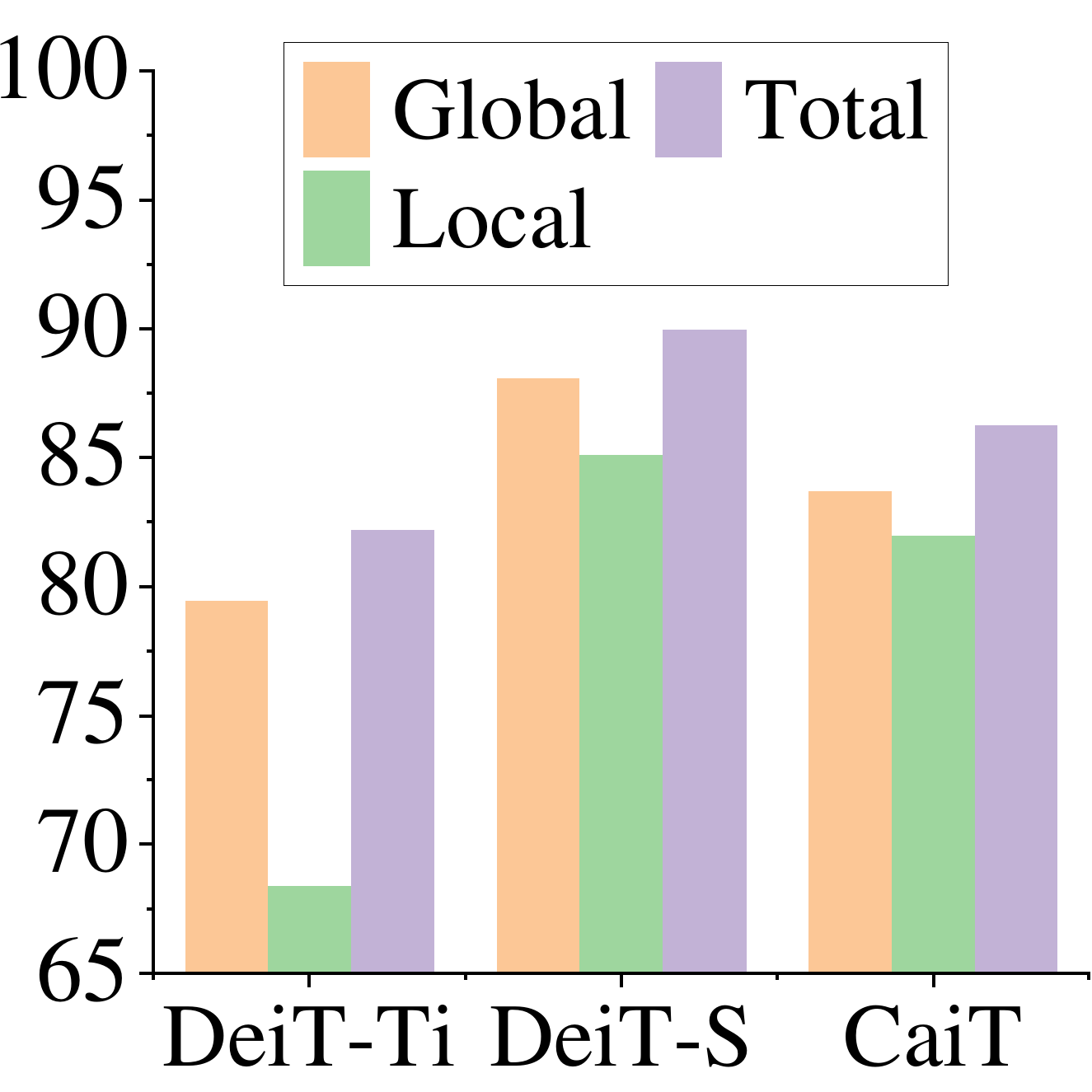}%
    }%
    \subcaptionbox{\normalsize {Cars}}{%
        \includegraphics[width=0.33\linewidth]{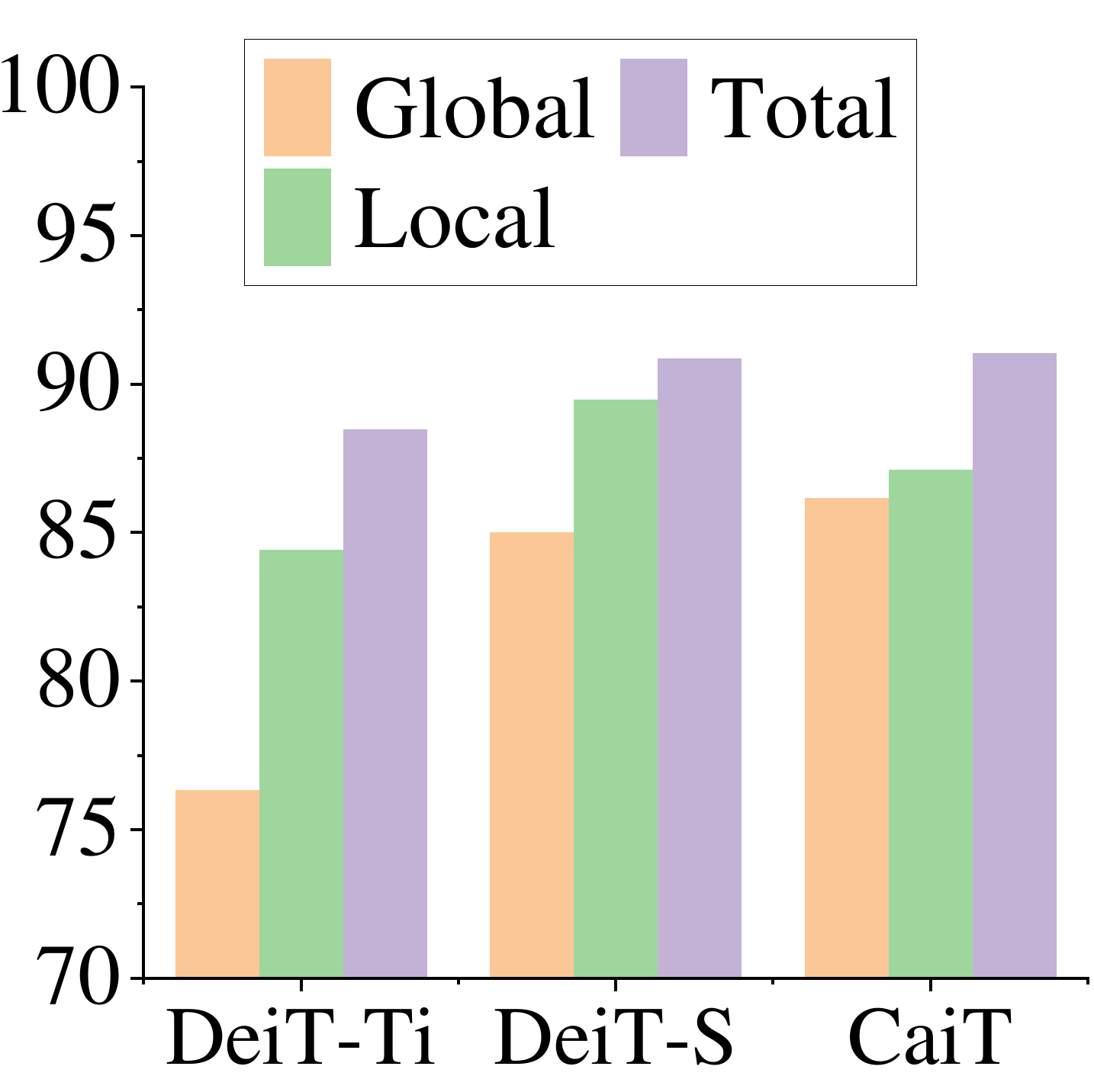}%
    }%
    \caption{Performance~(\%) of different branches (global branch, local and total) with three ViTs on adopted datasets.}
    \label{fig-6}  

\end{figure}

\begin{table}
\renewcommand\arraystretch{1.15}
\centering
\setlength{\tabcolsep}{0.6mm}{
\begin{tabular}{c|*4{c}|*4{c}}
  \toprule
\multirow{2}*{\textbf{Base}} & \multicolumn{4}{c|}{\textbf{Global Num.}}  & \multicolumn{4}{c}{\textbf{Local Num.}} \\

\cline{2-9}

& \textbf{3} & \textbf{5} & \textbf{10${}^{\dagger}$} & \textbf{15} & \textbf{5} & \textbf{10${}^{\dagger}$} & \textbf{15} & \textbf{20} \\

  \toprule
DeiT-Ti & 80.31 & 81.58 & 82.26 & 82.25 & 81.79 & 82.26 & 82.07 & 82.38 \\
DeiT-S & 83.40 & 84.50 & 84.85 & 84.90 & 84.39 & 84.85 & 84.93 & 85.16 \\

  \bottomrule
  \end{tabular}}
\caption{Results~(\%) of ProtoPFormer with different numbers of global and local prototypes per class on CUB.}
\label{tab-2}

\end{table}

\vspace{-1em}

\subsection{Ablation Study}


Specific accuracies of global and local branches are presented in Figure~\ref{fig-6} along with the combined total accuracy. Joint decisions achieve more promising results than separate global and local branches, verifying that the proposed global and local prototypes mutually correct each other with their perspective knowledge and benefit the whole model through the complementary decisions. Moreover, the influence of different numbers of global and local prototypes is also evaluated, as displayed in Table~\ref{tab-2}. The accuracy generally improves with the introduction of more prototypes, which also leads to parameter increase and high similarity between prototypes. Generally we choose medium numbers of prototypes in our experiment~(marked with $\dagger$). More ablation studies~(like the influence of used hyper-parameters) are presented in the supplementary material.






\section{Conclusion}
In this paper, we propose ProtoPFormer for appropriately and effectively applying the prototype-based method with ViTs for interpretable image recognition.
To fully capitalize on the built-in architectural characteristics of ViTs, ProtoPFormer introduces global and local prototypes, representing holistic and partial features respectively, which contribute associatively when inference.
In particular, the local prototypes are further supervised to tackle the ``prototype distraction'' problem by learning to concentrate on the visual pars individually.
Experiments have demonstrated the superiority of our ProtoPFormer.
In future work, we plan to investigate the potential applications of ProtoPFomer in other critical areas like model debugging and medical image diagnosis.

{\small
\bibliographystyle{ieee_fullname}
\bibliography{egbib}
}
\clearpage
\appendix
\section{Detailed Description of ProtoPNet}
We additionally describe ProtoPNet~\cite{chen2019looks-protopnet} method to complement the Preliminaries section in the main text. ProtoPNet uses an add-on layer to transform the output embedding features to the same dimension $d$ with prototypes. In our method, we fuse the vision transformer~(ViT) backbone and the add-on layer, termed as $\mathrm{Net}$. In ProtoPNet, the add-on layer contains two $1\times 1$ convolution layers and a sigmoid activation function. In our implementation, only one $1\times 1$ convolution layer and a sigmoid function make up the add-on layer. Moreover, ProtoPNet also introduces two new losses, and its overall optimization object is 
\begin{equation}
    \mathcal{L} = \mathcal{L}_{\mathrm{CE}} + \lambda_1 \mathcal{L}_{\mathrm{CLS}} + \lambda_2 \mathcal{L}_{\mathrm{SEP}} \text{,}
\end{equation}
where $\mathcal{L}_{\mathrm{CE}}$ is the conventional cross-entropy loss for penalizing the misclassified output logits from the fully-connected~(FC) layer with the ground-truth labels. 
And $\mathcal{L}_{\mathrm{CLS}}$ and $\mathcal{L}_{\mathrm{SEP}}$ are the cluster and separation loss multiplied with corresponding coefficients $\lambda_1$ and $\lambda_2$, separately. Let $\mathcal{T}=\{(I_i,y_i)\}^{N_T}_{i=1}$ be the training set, $\mathcal{L}_{\mathrm{CLS}}$ and $\mathcal{L}_{\mathrm{SEP}}$ can be defined as
\begin{equation}
    \mathcal{L}_{\mathrm{CLS}} = \frac{1}{N_T}\sum_{(I,y)\in \mathcal{T}}  \mathop{\text{min}}\limits_{j:p^{(j)}\in\mathcal{P}_y} \mathop{\text{min}}\limits_{\Tilde{I}\in \text{patches} (\mathrm{Net}(I))}
    \| \Tilde{I} - p^{(j)} \|^2_2 \text{,}
\end{equation}
and
\begin{equation}
    \mathcal{L}_{\mathrm{SEP}} = -\frac{1}{N_T}\sum_{(I,y)\in \mathcal{T}}  \mathop{\text{min}}\limits_{j:p^{(j)}\notin\mathcal{P}_y} \mathop{\text{min}}\limits_{\Tilde{I}\in \text{patches} (\mathrm{Net}(I)))}
    \| \Tilde{I} - p^{(j)} \|^2_2 \text{.}
\end{equation}
$\Tilde{I}$ denotes the latent features of the transformed input $I$ and $\mathcal{P}_y$ represents the prototype set belonging to the class $y$.
These two losses enforce the prototypes learn to be activated with their task-specific inputs and widen the inter-class feature-prototype distances.
In our ProtoPFormer, the proposed prototypical part concentration~(PPC) loss, $\mathcal{L}_{\mathrm{PPC}}$, is effective enough to promote the divergency of prototypes and learn diverse discriminative features of visual evidence. For this reason, we do not employ $\mathcal{L}_{\mathrm{CLS}}$ and $\mathcal{L}_{\mathrm{SEP}}$ in our method and the final loss of ProtoPFormer is $\mathcal{L} = \mathcal{L}_{\mathrm{CE}} + \mathcal{L}_{\mathrm{PPC}}$, as stated in the main text.

A ``push'' process is also adopted in ProtoPNet by projecting the prototypes onto the closest patches every five epochs in the training stage to ensure that the prototypes always equal some latent features of training image patches. Our proposed ProtoPFormer does not employ the ``push'' process for two main reasons. One is that this process causes the performance degradation with ViT backbones, and the other is that our global and local prototypes are the high-level abstraction of associated visual explanations, representing their features based on the whole training set.

For the FC layer, $m$ similarity scores are passed through a fully connected layer, parameterized by $\Omega \in \mathbb{R}^{m}$, to produce the logits for classification. Moreover, for a class $c$ and prototype $p^{(i)}$, if $p^{(i)}$ belongs to class $c$, $\Omega_{c,i}$ is set to $1$, otherwise it's set to $-0.5$. 
In our ProtoPFormer, the FC layers are non-trainable in global and local branches. Hence, in the reasoning process of ProtoPFormer shown in Figure~\ref{fig-3} in the main paper, the class connection row is omitted as the connection weight is permanently fixed as $1$~(connecting prototypes to the logit of its class) and $-0.5$~(connecting prototypes to the logit of the other class).

\section{Experimental Details}

We give more implementation details of our experiments in this section.

\subsection{Datasets}

We give the specific numbers of training and test images and classes of adopted CUB~\cite{399-cub}, Dogs\cite{KhoslaYaoJayadevaprakashFeiFei_FGVC2011-dogs} and Cars~\cite{krause20133d-stanfordcars} datasets in Table~\ref{tab:datasets}. As stated in the main text, we use full images~(un-cropped) in our experiments, unlike  ProtoPNet~\cite{chen2019looks-protopnet}.



\begin{table}[h]
\centering
\setlength{\tabcolsep}{9pt}
\begin{tabular}{l rrr}
    \toprule
    \multicolumn{1}{c}{\textbf{dataset}}
    & \multicolumn{1}{c}{\textbf{Train}}
    & \multicolumn{1}{c}{\textbf{Test}}
    & \multicolumn{1}{c}{\textbf{Classes}} \\
    \midrule
    CUB & 5994 & 5794 & 200 \\
    Dogs & 12000 & 8580 & 120 \\
    Cars & 8144 & 8041 & 196 \\
    \bottomrule
\end{tabular}
\caption{Detailed statistics of adopted three datasets.}
\label{tab:datasets}
\end{table}

\subsection{Experimental Settings}

\paragraph*{Training Settings.}
In this subsection, our experimental setups are depicted in detail. When training, we basically follow the training strategies used in official codes~\footnote{https://github.com/facebookresearch/deit} of DeiT~\cite{touvron2021deit} in PyTorch~\cite{NEURIPS2019_9015-pytorch} on four Nvidia Tesla A100 GPUs. All input images are resized to $224\times 224$ pixels before feeding into used ViT backbones that are initialized with the official pre-trained weights on ImageNet-1k~\cite{deng2009imagenet}. The dimension $d$ of prototypes is set to $192$ for all methods, which include global and local prototypes in our ProtoPFormer and prototypes in prototype-based baselines.
We use the AdamW~\cite{loshchilov2017decoupled-adamw} optimizer with different learning rates using a cosine annealing scheduler~\cite{loshchilov2016sgdr-cosine}.  
All the models are trained for $200$ epochs~($5$ warm-up epochs with a learning rate $1e^{-4}$) with a batch size of $128$. After warming up, the learning rates of the add-on layer and prototypes are turned to $3e^{-3}$ and $3e^{-3}$, respectively. For the ViT backbones, different learn rates are employed on different datasets, \ie, $1e^{-4}$ for CUB, $5e^{-5}$ for Dogs and $4e^{-4}$ for Cars, respectively. Many data augmentation techniques are adopted in our experiments following the settings in official codes of DeiT, including label smoothing~\cite{szegedy2016rethinking-label_smoothing}, stochastic depth~\cite{huang2016deep-stochastic_depth}, Mixup\cite{zhang2017mixup-mixup}, Cutmix\cite{yun2019cutmix-cutmix} repeated augmentation~\cite{berman2019multigrain-repeat_aug,hoffer2020augment-repeat_aug}, Random Erasing~\cite{zhong2020random-random_earsing} and RandAugment~\cite{cubuk2020randaugment-randaugment}. Since the DeiT-style settings have been widely employed in many ViT-backed methods, our proposed ProtoPFormer can benefit and boost the applications of prototype-based methods in ViT-related areas and approaches.

For a fair comparison, we retrain all the baselines with adopted three ViTs under the same training settings with ProtoPFormer except for the learning rates. We find that different baselines achieve optimal performance with different learning rates. The grid search strategy is adopted to find the optimal learning rates of each baseline. Please see the following section for specific learning rates.
Then we report each method's performance averaged on six runs with six seeds~(1028, 2678, 3566, 4686, 5328, and 6186, separately), given in Table~\ref{tab-1} in the main paper.




\paragraph*{Hyperparameters of Baselines}
Here, we list the involved hyperparameters of baseline methods. We use grid search to find the optimum values~(like learning rates and their exclusive parameters) of compared baselines when applying them with ViT backbones.
\begin{itemize}
    \item \textbf{Base} As the non-interpretable counterpart of ProtoPFormer, we use the official implementation of three ViT backbones, \ie, DeiT-Ti~\cite{touvron2021deit}, DeiT-S~\cite{touvron2021deit}, and CaiT-XXS-24~\cite{touvron2021going-cait}. After warming up (5 warm-up epochs with a learning rate $1e^{-4}$), $5e^{-5}$, $1e^{-4}$ and $4e^{-4}$ are learning rates of backbones on CUB, Dogs and Cars, respectively.
    \item \textbf{ProtoPNet}~\cite{chen2019looks-protopnet} We use ten prototypes for each class on the adopted three datasets. In our implemented ProtoPNet, the add-on layer only consists of one convolutional layer, and the last layer of the model is frozen for better results. Some operations in the original ProtoPNet are also removed for performance improvement with ViT backbones, including the cluster and separation losses and the ``push'' process. After warming up (5 warm-up epochs with a learning rate $1e^{-4}$), the learning rates of the add-on layer and the prototypes are set to $3e^{-3}$. And $2e^{-4}$, $5e^{-5}$ and $2e^{-4}$ are learning rates of backbones for CUB, Dogs and Cars, respectively.
    \item \textbf{ProtoTree}~\cite{nauta2021neural-prototree} For the soft decision tree, we use a single tree with a depth of 12 with 4095 total prototypes for classification (depth 9 is used in the original ProtoTree). After warming up (5 warm-up epochs with a learning rate $1e^{-4}$), the learning rates of the add-on layer and the prototypes are turned to $3e^{-3}$. And $1e^{-4}$, $5e^{-5}$ and $4e^{-4}$ are learning rates of ViT backbones for CUB, Dogs and Cars, respectively.
    \item \textbf{TesNet}~\cite{wang2021interpretable-proto_TesNet} Each class contains ten prototypes on three datasets. After warming up (5 warm-up epochs with a learning rate $1e^{-4}$), the learning rates of the add-on layer, the prototypes, and the last layer are turned to $3e^{-3}$, $3e^{-3}$ and $1e^{-4}$, respectively. And $1e^{-4}$, $1e^{-4}$ and $2e^{-4}$ are learning rates of backbones for CUB, Dogs and Cars, respectively. The cluster and separation loss, and the ``push'' process are also removed in this work for better performance.
    \item \textbf{Def.ProtoPNet}~\cite{donnelly2022deformable-protopnet} We use ten deformable prototypes with the shape of $2\times 2$ for each class on the adopted three datasets ($2\times 2$ is the same as the defaulting setting in the released codes). The add-on layer of the original Def. ProtoPNet is an upsampling layer to increase the feature map size of CNN models. We replace the upsampling layer with a $1\times 1$ convolution layer and an activation layer, which improves the accuracy in our experiments. After warming up (5 warm-up epochs with a learning rate $1e^{-4}$), the learning rates of the add-on layer, the prototypes, and the last layer are set to $3e^{-3}$, $3e^{-3}$ and $1e^{-4}$, respectively. And $5e^{-5}$, $2e^{-4}$ and $4e^{-4}$ are learning rates of backbones for CUB, Dogs and Cars, respectively.
    \item \textbf{CT}~\cite{rigotti2021attention-ConceptTransformer} We only conduct experiments with CT on the CUB dataset. 13 concepts,  95 spatial concepts, and  0 unsupervised concepts are adopted in this method. The learning rate of the whole model is $1e^{-4}$.
    \item \textbf{ViT-Net}~\cite{kim2022vit-vit_net} We use a single tree with a depth of 4 with 15 prototypes in ViT-Net. After warming up (5 warm-up epochs with a learning rate $1e^{-4}$), the learning rates of the whole model is turned to $1e^{-4}$, $5e^{-5}$ and $2e^{-4}$ on CUB, Dogs, and Cars, respectively.
\end{itemize}

Especially, CT and ViT-Net are designed based on ViTs. CT uses ViT-L\cite{dosovitskiy2020vit} in the original paper, and ViT-Net unitizes the pre-trained weights on ImageNet-22K~\cite{deng2009imagenet} rather than ImageNet-1k used in our settings. We turn their hyperparameters for fairness and find suitable configurations under our experimental settings. 
It is worth noticing that we modify some modules in some baselines. For example, the add-on layer of ProtoPNet is changed to one convolutional layer in our implementation. These changes are implemented in order to pursue better accuracies and visualization results for compared baselines with ViT backbones.





\section{Additional Ablation Study}

We present qualitative analyses for involved hyperparameters and more discussions in this section.


\subsection{Influence of Kept Tokens}
\begin{figure}[htb]
\centering
    \includegraphics[width=\linewidth]{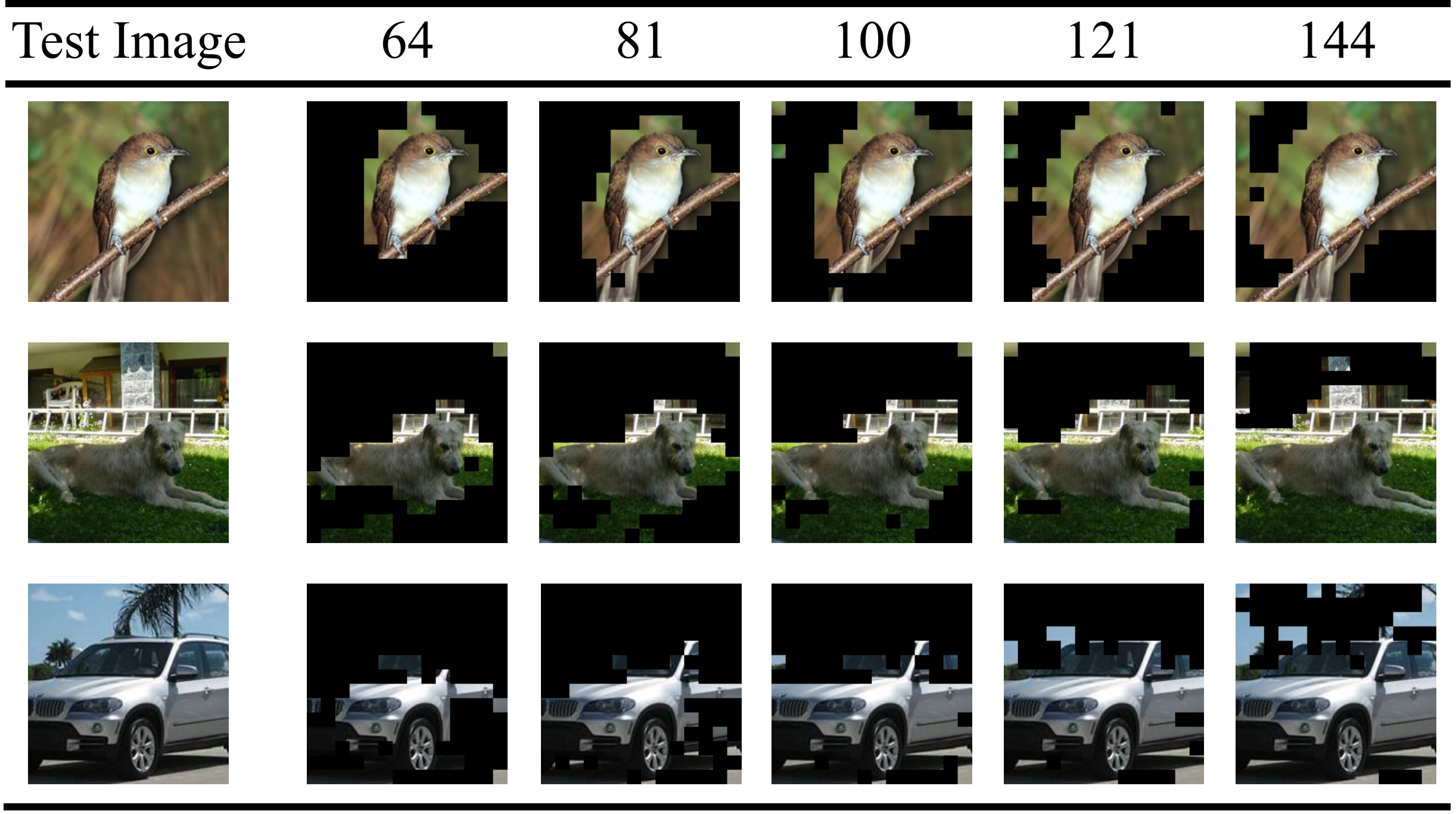}
\caption{Visualization comparison of different numbers of preserved image tokens on three test images randomly chosen from the CUB, Dogs, and Cars datasets.}
\label{fig-3-mask}
\end{figure}

We compare different numbers $K$ of kept foreground-related image tokens in Table~\ref{tab:mask}. The illustration of the foreground preserving~(FP) masks with different $K$ are given in Figure~\ref{fig-3-mask}. 
For CUB and Dogs datasets, a small value of $K$ can preciously screen the background-related image tokens because their target objects, \ie, birds and dogs, are prone to appear in a relatively small area in the whole image. While cars, in the Cars dataset, tend to occupy most of the entire screen of the pictures. Therefore, we set $K$ to 81 for CUB and Dogs datasets and a more significant value, 121, for the Cars dataset with a total $196$ image token numbers.

\begin{table}
\renewcommand\arraystretch{1.15}
\centering

\begin{tabular}{c|*5{c}}
  \toprule
\textbf{Dataset} & \textbf{64} & \textbf{81} & \textbf{100} & \textbf{121} & \textbf{144}  \\

  \toprule
CUB & 81.91 & 82.26 & 82.05 & 82.07 & 82.08 \\
Dogs & 81.86 & 82.20 & 81.93 & 82.03 & 82.12 \\
Cars & 87.85 & 87.95 & 88.10 & 88.48 & 88.36 \\

  \bottomrule
\end{tabular}

\caption{Results~(\%) of ProtoPFormer with different numbers of kept image tokens with DeiT-Ti.}
\label{tab:mask}

\end{table}

\subsection{Influence of Weighted
Coefficients}
The influence of two weighted coefficients $\lambda_{\mu}$ and $\lambda_{\sigma}$ is analyzed in Table~\ref{tab-coefficients}.
In this table, we compare different values of these two coefficients to find the best final decisions jointly made by the global and local branches. We adopt the optimal pairs of $\lambda_{\mu}=0.5$ and $\lambda_{\sigma}=1$  for our method's mutual correction and joint decision.


\begin{table}[h]
\renewcommand\arraystretch{1.15}
\centering

\begin{subtable}[h]{1.0\linewidth}
\centering
\begin{tabular}{c|*4{c}}
  \toprule
\multirow{2}*{\textbf{Base}} & \multicolumn{4}{c}{\textbf{$\lambda_{\mu}$}}   \\
\cline{2-5}
& \textbf{0.1} & \textbf{0.3} & \textbf{0.5${}^{\dagger}$} & \textbf{0.7}  \\
  \toprule
DeiT-Ti & 82.14 & 82.31 & 82.26 & 81.96 \\
DeiT-S & 84.98 & 84.75 & 84.85 & 84.79 \\
  \bottomrule
\end{tabular}
\caption{Influence of $\lambda_{\mu}$ with $\lambda_{\sigma}=0.1$}
\label{tab:coefficients-of-mean}
\end{subtable}

\begin{subtable}[h]{\linewidth}
\centering
\begin{tabular}{c|*4{c}}
  \toprule
\multirow{2}*{\textbf{Base}} & \multicolumn{4}{c}{\textbf{$\lambda_{\sigma}$}}   \\
\cline{2-5}
& \textbf{0.04} & \textbf{0.1${}^{\dagger}$} & \textbf{0.2} & \textbf{0.5}  \\
  \toprule
DeiT-Ti & 82.12 & 82.26 & 81.94 & 81.83 \\
DeiT-S & 84.97 & 84.85 & 84.86 & 84.73 \\
  \bottomrule
\end{tabular}
\caption{Influence of $\lambda_{\sigma}$ with $\lambda_{\mu}=0.5$}
\label{tab:coefficients-of-sigma}
\end{subtable}

\caption{Results~(\%) of ProtoPFormer with different values of two weighted coefficients $\lambda_{\mu}$ and $\lambda_{\sigma}$ on the CUB dataset. Values of our adopted coefficients are marked with $\dagger$.}
\label{tab-coefficients}
\end{table}

\subsection{Influence of Thresholds}
The choice two thresholds $t_{\mu}$ and $t_{\sigma}$
also affects the performance of ProtoPFormer. The experiments of different thresholds are conducted and presented in Table~\ref{tab-thresholds}. The best pair of these two thresholds, $t_{\mu}=2$ and $t_{\sigma}=1$, is chosen in our ProtoPFormer.


\begin{table}
\renewcommand\arraystretch{1.15}
\centering

\begin{subtable}[t]{\linewidth}
\centering
\begin{tabular}{c|*5{c}}
  \toprule
\multirow{2}*{\textbf{Base}} & \multicolumn{5}{c}{\textbf{$t_{\mu}$}}   \\
\cline{2-6}
& \textbf{0.5} & \textbf{1.0} & \textbf{1.5} & \textbf{2.0${}^{\dagger}$} & \textbf{2.5}  \\
  \toprule
DeiT-Ti & 82.00 & 82.07 & 82.15 & 82.26 & 81.93 \\
DeiT-S & 84.83 & 84.93 & 84.94 & 84.85 & 85.09 \\
  \bottomrule
\end{tabular}
\caption{Influence of $t_{\mu}$ with $t_{\sigma}=1$}
\label{tab:coefficients-of-sigma}
\end{subtable}

\begin{subtable}[t]{\linewidth}
\centering
\begin{tabular}{c|*5{c}}
  \toprule
\multirow{2}*{\textbf{Base}} & \multicolumn{5}{c}{\textbf{$t_{\sigma}$}}   \\
\cline{2-6}
& \textbf{0.5} & \textbf{1.0${}^{\dagger}$} & \textbf{2.0} & \textbf{4.0} & \textbf{8.0} \\
  \toprule
DeiT-Ti & 82.02 & 82.26 & 82.21 & 82.29 & 82.19 \\
DeiT-S & 85.09 & 84.85 & 84.97 & 85.17 & 84.97 \\
  \bottomrule
\end{tabular}
\caption{Influence of $t_{\sigma}$ with $t_{\mu}=2$}
\label{tab:coefficients-of-mean}
\end{subtable}

\caption{Results~(\%) of ProtoPFormer with different values of two thresholds $t_{\mu}$ and $t_{\sigma}$ on the CUB dataset, where $\dagger$ mark denotes the thresholds we used in our experiments.}
\label{tab-thresholds}
\end{table}

\subsection{Influence of Prototype Dimension}
We evaluate the performance of different prototype dimensions $d$, presented in Table~\ref{tab:prototypes-dimension}.
It can be noticed that both too small and too big dimensions impair the classification accuracy.
In addition, the embedding dimension of tokens remains consistent with that of prototypes, which implies that the too small and too big embedding dimensions also lead to information loss and the introduction of the noise into latent features individually.
Consequently, we choose a medium value, $192$, as the dimension of global and local prototypes in all experiments.




\begin{table}
\centering

\begin{tabular}{c|*4{c}}
  \toprule
\textbf{Base} & \textbf{128} & \textbf{192${}^{\dagger}$} & \textbf{384} & \textbf{512}  \\

  \toprule
DeiT-Ti & 82.22 & 82.26 & 82.05 & 81.93 \\
DeiT-S & 84.83 & 84.85 & 84.45 & 84.74  \\

  \bottomrule
\end{tabular}
\caption{Results~(\%) of ProtoPFormer with different prototypes dimensions $d$ on CUB, the $\dagger$ mark denotes the dimension we used.}

\label{tab:prototypes-dimension}

\end{table}

\section{Additional Visualization Analysis}

More visualization results are given in this section.

\subsection{Reasoning Process}

\begin{figure}[!t]
\centering

    \includegraphics[width=\linewidth]{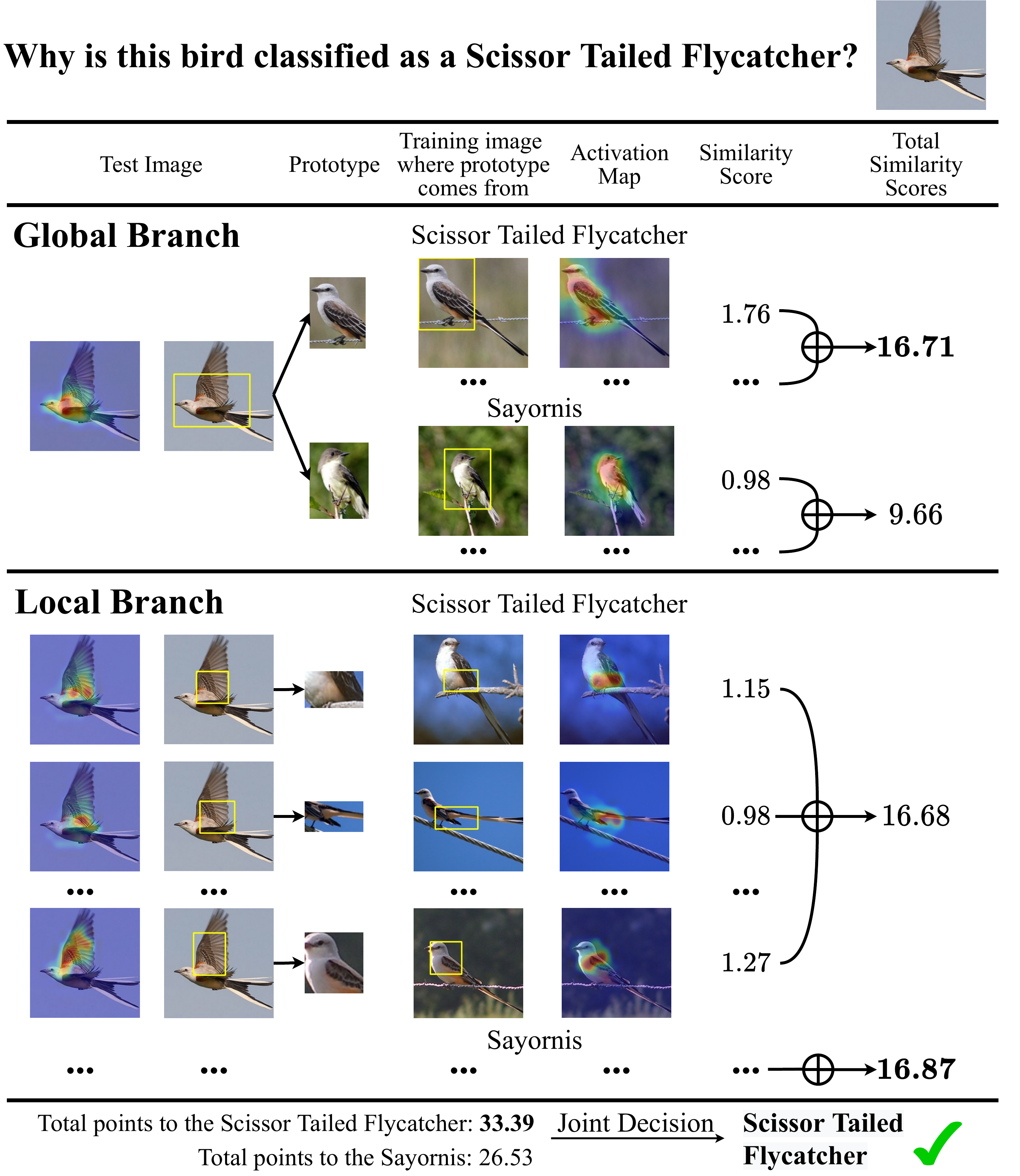}
\caption{The reasoning process of our ProtoPFomer in classifying the species of a bird with DeiT-Ti, where $\bm{\oplus}$ denotes summation of similarity scores.}
\label{fig-4-global-local-1}
\end{figure}

We additionally present two reasoning processes with the mutual correction and joint decision strategy of ProtoPFormer to better understand our method.
As demonstrated in Figure~\ref{fig-4-global-local-1}, the local branch makes the wrong prediction that a  Scissor Tailed Flycatcher is mistakenly classified as a Sayornis. Fortunately, the global branch makes the right decision by recognizing the informative holistic features of the target class. It corrects the local branch through the joint decision strategy, triumphantly pointing out visual evidence for interpretable image recognition.
Collectively, ProtoPFormer can faithfully reason the decision-making process of ViT backbones from the global and local view while achieving superior performance with the mutual decision and joint decision strategy of our proposed global and local prototypes.






\subsection{Comparison of Gaussian Function Shape}

\begin{figure*}[!t]
\centering

    \includegraphics[width=\linewidth]{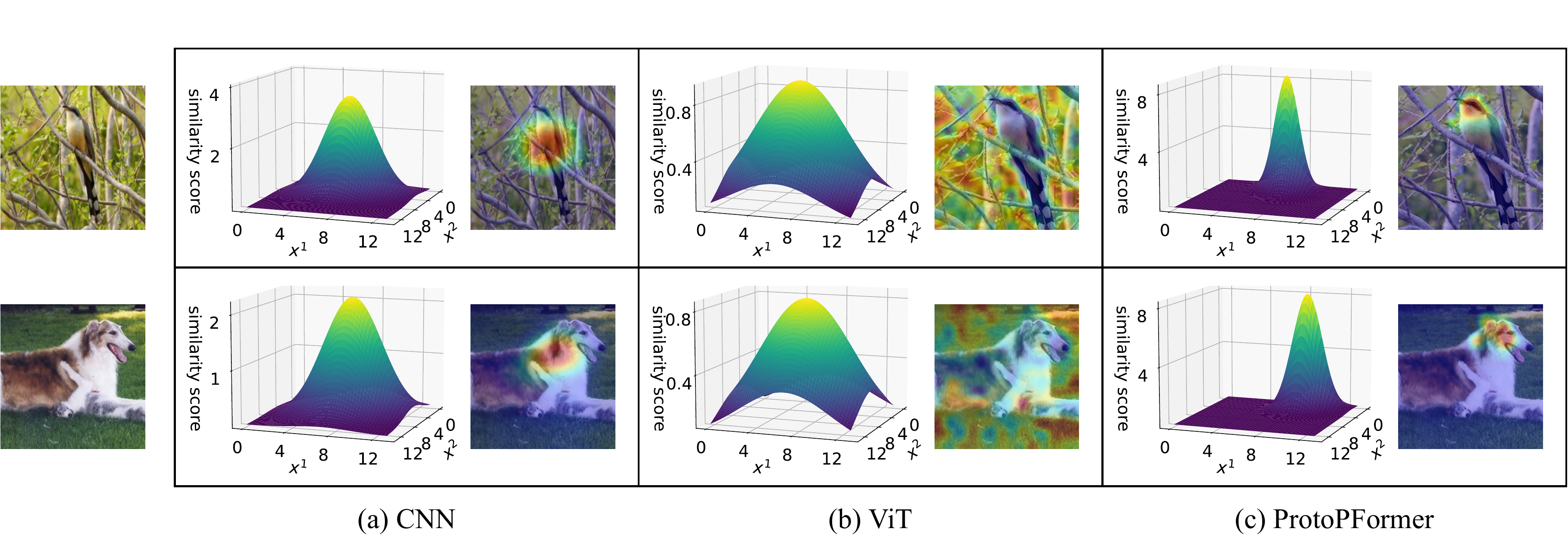}
\caption{The Gaussian function shape comparison of leaned prototypes~(in a heat map) between a CNN-based ProtoPNet (ResNet34), a ViT~based ProtoPNet~(DeiT-Ti) and our ProtoPFormer~(DeiT-Ti) with test images randomly chosen from CUB and Dogs.}
\label{fig-2-Gaussian}
\end{figure*}

We compare the shapes of the Gaussian function fitted by the activation scores of learned prototypes~(in a heat map) on test images between a CNN-based ProtoPNet~(ResNet34~\cite{he2016resnet}), a ViT~based ProtoPNet~(DeiT-Ti) and our ProtoPFormer~(DeiT-Ti), shown in Figure~\ref{fig-2-Gaussian}. 
It can be seen that the Gaussian function of the  CNN-based prototype maintains smaller covariance value than that of the ViT-based prototype. This phenomenon inspires us to propose the prototypical part concentration~(PPC) to penalize too close center coordinates and encourages small covariance values, described in the Method section in the main text. As shown in Figure~\ref{fig-2-Gaussian}~(c), with the PPC loss the local prototype can concentrate on diverse discriminative parts, thereby providing understandable visual explanations for interpretable image recognition.



\subsection{Visualization of Global Prototypes}
\begin{figure*}[t]
\centering

    \includegraphics[width=\linewidth]{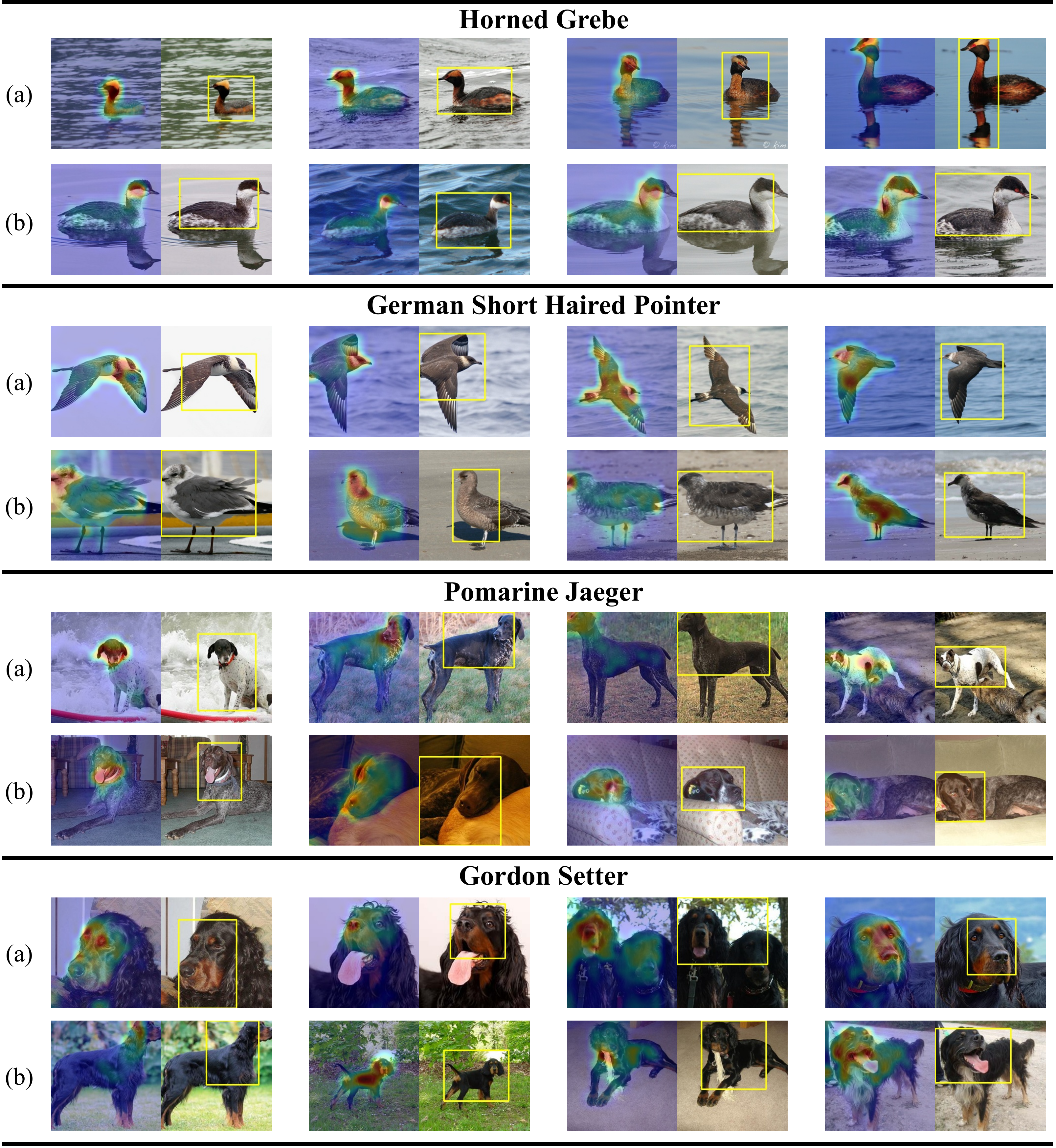}
\caption{Visualization of different global prototypes of four categories of on the CUB and Dogs datasets with DeiT-Ti. The selected examples have the highest activation scores with each global prototype.}
\label{fig-5-global-proto}
\end{figure*}

The visualization of two global prototypes (a) and (b) on sample images is shown in Figure~\ref{fig-5-global-proto}.
We can observe that different global prototypes mainly identify different kinds of holistic features of targets. For instance, two global prototypes of the German Short Haired Pointer learn the differences between a flying bird and a perching bird; and for Gordon Setter, two global prototypes learn the features of a dog's front and side views.
In summary, global prototypes mainly capture the
holistic features of targets with pose variations and different perspectives, providing the interpretability from the global view for ProtoPFormer.

\subsection{Visual Comparison}
\begin{figure*}[htb]
\centering
    \includegraphics[width=\linewidth]{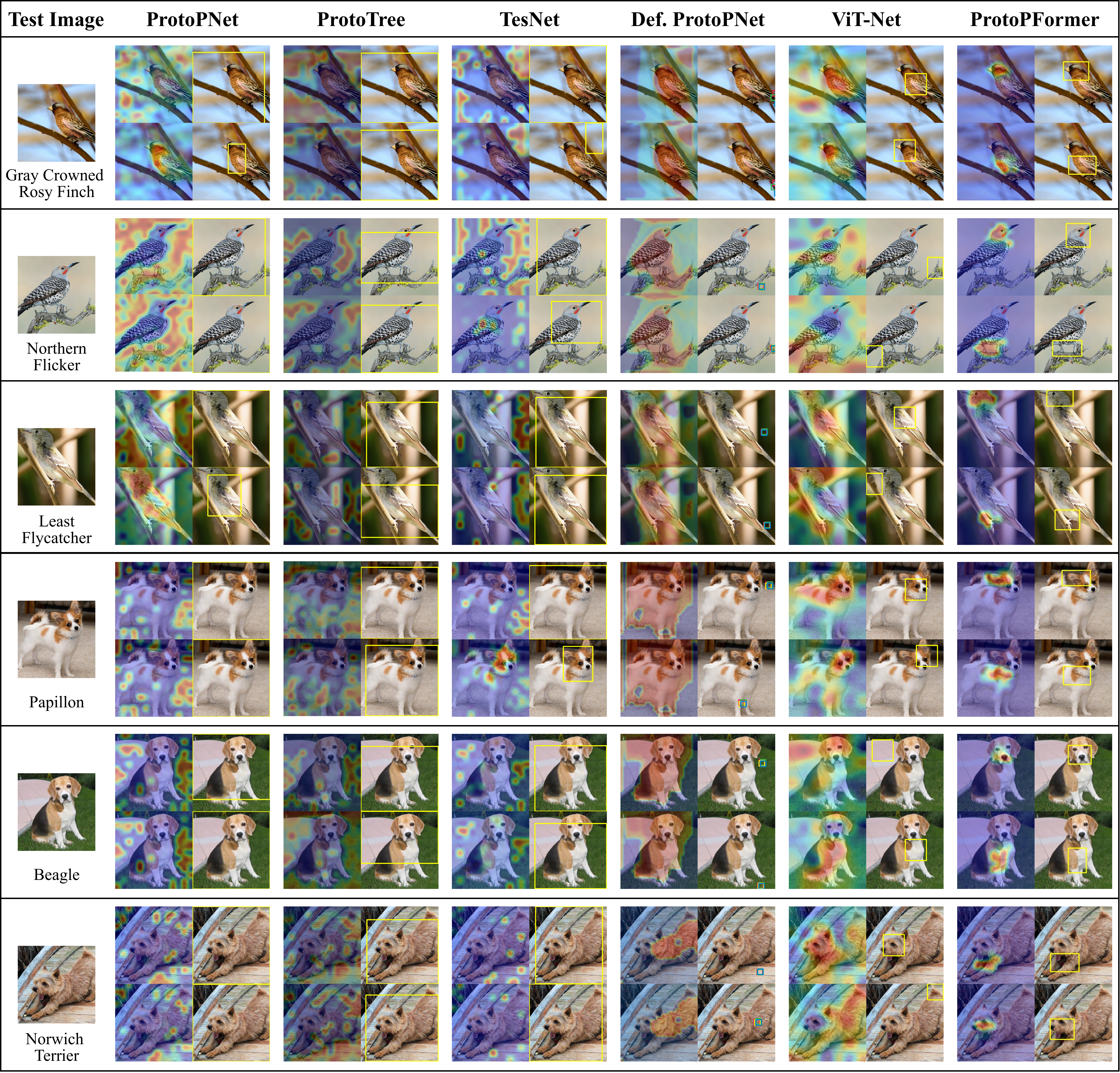}
\caption{Visual demonstration of the two most activated local prototypes in heat maps and bounding boxes on example images~(randomly chosen from the CUB and Dogs datasets) of five prototype-based baselines and ProtoPFormer with DeiT-S.}
\label{fig-1-cub-dog}
\end{figure*}

\begin{figure*}[!t]
\centering
    \includegraphics[width=\linewidth]{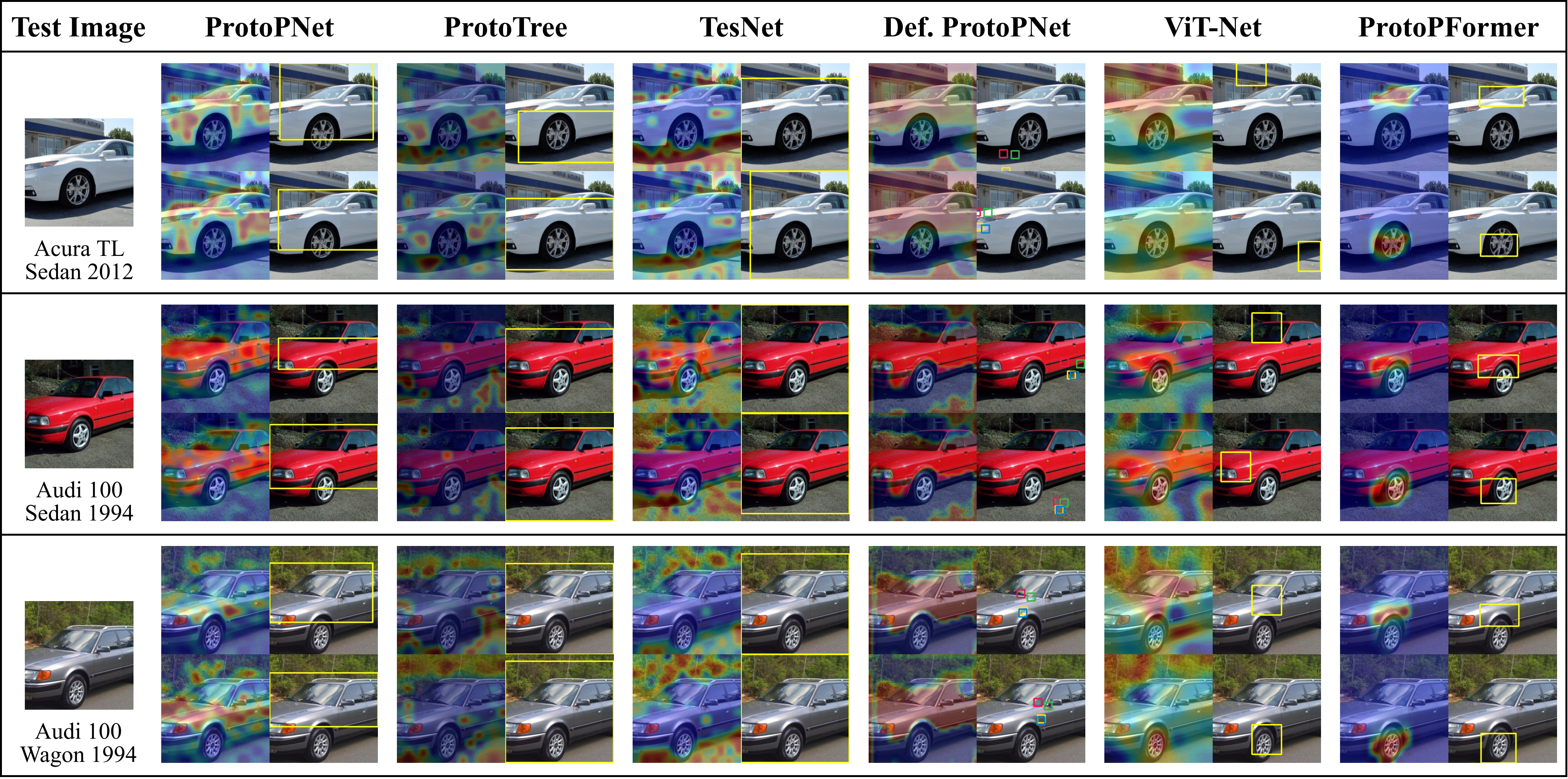}
\caption{Visual demonstration of the two most activated local prototypes in heat maps and bounding boxes on example images~(randomly chosen from the Cars dataset) of five prototype-based baselines and ProtoPFormer with DeiT-S.}
\label{fig-1-car}
\end{figure*}

More visual comparisons between five prototype-based baselines and our ProtoPFormer on the CUB, Dogs, and Cars datasets, are depicted in Figure~\ref{fig-1-cub-dog} and Figure~\ref{fig-1-car}. The comparison results are consistent with the main paper. Five baseline methods show scattered activations on the test images, failing to find visual evidence for interpretable reasoning of the ViT backbone.
By contrast, the proposed ProtoPFormer has achieved the best visualization performance with the ViT backbone, which successfully highlights discriminative parts of corresponding categories, like, the head and feet of birds, the head and belly of dogs, and the window and wheel of cars.

\end{document}